\pdfoutput=1
\documentclass[11pt,onecolumn]{cleantechnicalreport}

\usepackage[numbers,sort&compress,square]{natbib}
\usepackage{hybridfrontmatter}
\usepackage{multirow,makecell,siunitx,adjustbox,wrapfig}
\usepackage{algorithm,algorithmic}

\newenvironment{compactmath}{%
    \small
    \setlength{\baselineskip}{12pt}%
    \setlength{\abovedisplayskip}{5pt plus 2pt minus 0.5pt}%
    \setlength{\belowdisplayskip}{5pt plus 2pt minus 0.5pt}%
    \setlength{\abovedisplayshortskip}{0pt plus 2pt}%
    \setlength{\belowdisplayshortskip}{2pt plus 2pt}%
}{\ignorespacesafterend}

\definecolor{linkblue}{HTML}{2456A6}
\hypersetup{allcolors=linkblue,breaklinks=true}

\title{\textcolor{ReportAccent}{\textbf{\textit{VimRobo}}}: Empowering Spatio-Temporal Reasoning over Long-Horizon Embodied Experience via In-Context Scene Relation Reconstruction}

\reportauthors{Qiuchen Wang$^{1,2}$, Shunyu Liu$^{1,\dagger}$, Junjie Zhang$^{1}$,\\
  Yilun Kong$^{1}$, Zheng Zhang$^{1}$, Dacheng Tao$^{1,\ddagger}$}
\reportaffiliation{$^{1}$Nanyang Technological University \quad $^{2}$ACE Robotics}
\reportauthornote{$^{\dagger}$ Project Leader. \quad $^{\ddagger}$ Corresponding Author.}
\reportlogo[3cm]{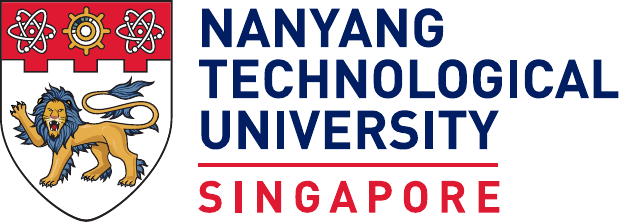}
\reportpartnerlogo[2.4cm]{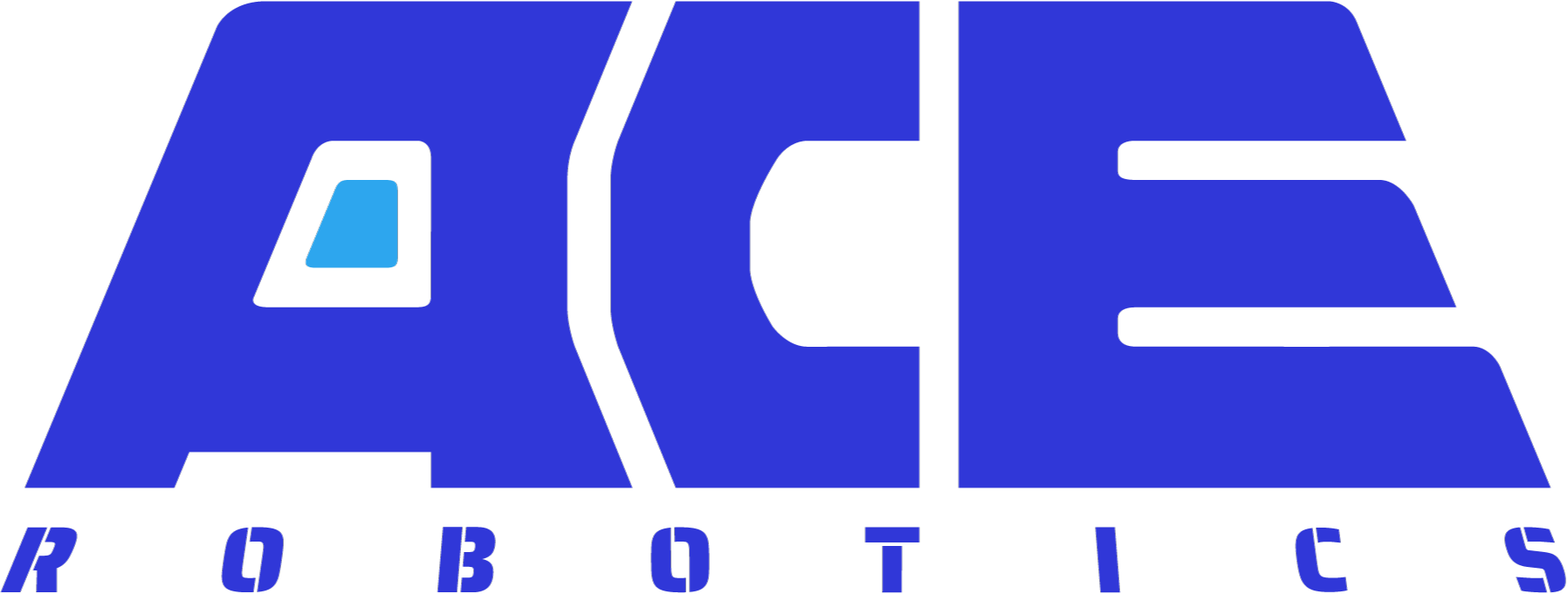}
\reportheader{\textcolor{ReportTheme}{\textbf{GrAIL}}\hspace{0.6em}\textcolor{gray}{\rule[-0.25ex]{0.4pt}{1.1em}}\hspace{0.6em}\textcolor{black!60}{\textsc{Nanyang Technological University}}}

\reportmeta{\faEnvelope}{Contact}{qiuchen.wang@ntu.edu.sg}
\reportmeta{\faGlobe}{Project Page}{\href{https://atomrecur.github.io/VimRobo/}{\nolinkurl{https://atomrecur.github.io/VimRobo/}}}
\reportmeta{\faGithub}{Github}{\href{https://github.com/AtomRecur/VimRobo}{\nolinkurl{https://github.com/AtomRecur/VimRobo}}}
\reportmeta{\faCube}{Models}{\href{https://huggingface.co/collections/AtomRecur/vimrobo}{\nolinkurl{https://huggingface.co/collections/AtomRecur/vimrobo}}}

\DeclareCaptionFont{capsize}{\fontsize{9.5}{11.5}\selectfont}
\makeatletter
\providecommand{\bstctlcite}[1]{\@bsphack\immediate\write\@auxout{\string\citation{#1}}\@esphack}
\makeatother

\begin{abstract}
Reasoning over long-horizon embodied experience requires models to understand how visual evidence relates to motion, location, and time. However, existing methods reconstruct scenes or build explicit scene graphs offline, providing limited flexibility for organizing task-relevant evidence across instances within the model's context. We introduce \mbox{\textbf{VimRobo}} for online, in-context reconstruction of task-relevant scene relations. Our pilot study examines how memory representations and retrieval perspectives influence evidence localization and end-to-end accuracy. Building on these findings, we propose visual and motion memory from embodied experience and design multi-perspective memory retrieval to distinguish semantically similar observations using spatial or temporal task anchors. We further introduce in-context scene relation reconstruction, using task-specific skills to organize and visually verify retrieved evidence within this memory, shaping the agent's context for reasoning about scene relations. For training and evaluation, we develop a Q\&A synthesis pipeline and construct RoboMemBench to assess cross-instance reasoning. Experiments on various embodied memory benchmarks demonstrate gains over baselines in descriptive, spatial, and temporal reasoning.
\end{abstract}

\begin{document}
\bstctlcite{IEEEexample:BSTcontrol}

\maketitle
\suppressfloats[t]
\begin{figure}[h]
    \centering
    \vspace{-20pt}
    \includegraphics[width=\textwidth]{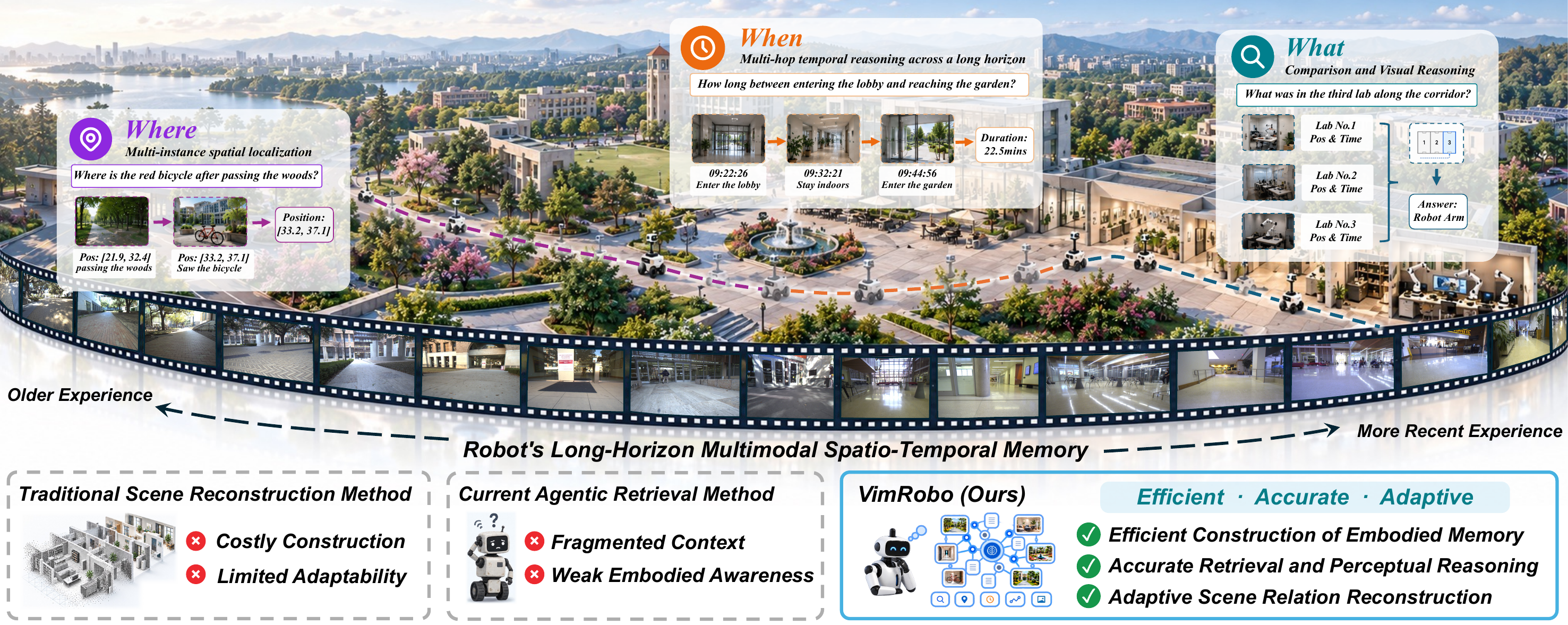}
    \vspace{-20pt}
    \caption{VimRobo is designed for temporal, spatial, and descriptive reasoning over long-horizon embodied experience. Traditional scene reconstruction is costly and rigid, while agentic retrieval suffers from fragmented context and weak embodied awareness. We combine efficient visual and motion memory construction, accurate multi-perspective retrieval, and adaptive in-context scene relation reconstruction.}
    \label{fig:teaser}
\end{figure}

\begin{figure}[!t]
    \centering
    \includegraphics[width=\textwidth]{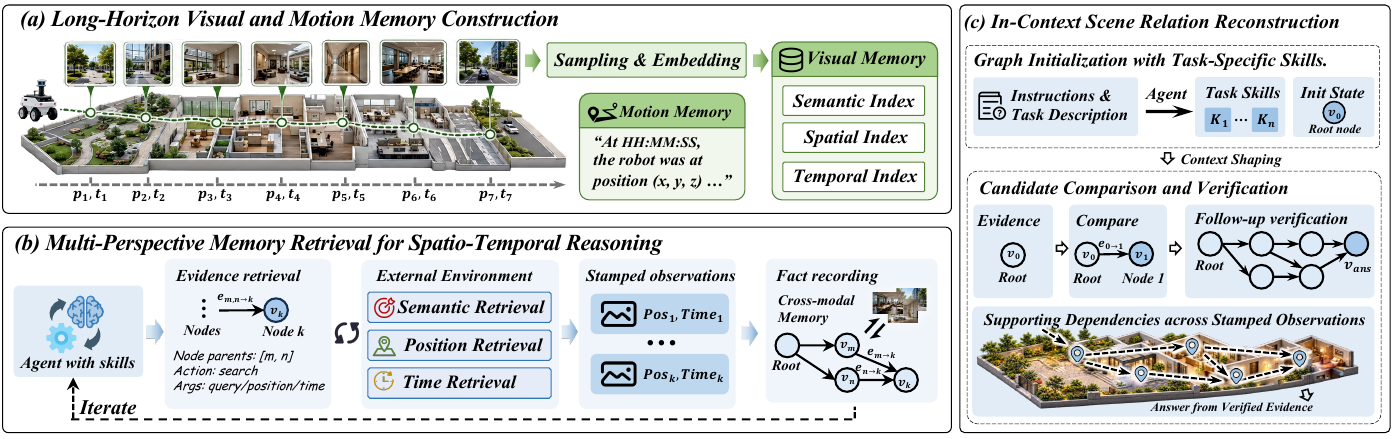}
    \vspace{-17pt}
    \caption{Overview of the VimRobo framework for embodied reasoning.
    (a)~Visual and Motion Memory Construction indexes evidence and summarizes motion.
    (b)~Multi-Perspective Memory Retrieval locates evidence and reranks semantic candidates by spatial or temporal anchors.
    (c)~In-Context Scene Relation Reconstruction uses task-specific skills to organize and visually verify evidence for scene relations.}
    \label{fig:framework}
    \vspace{-10pt}
\end{figure}

\section{Introduction}
\label{sec:introduction}
Recent Multimodal Large Language Models (MLLMs) advance real-world spatio-temporal reasoning, supporting embodied applications such as scene perception and spatial relation understanding~\citep{Qwen3-VL,chen2024spatialvlm,cheng2024spatialrgpt,ma2025spatialllm,zhu2025llava3d,gu2024conceptgraphs,werby2024hovsg,maggio2024clio}. In practice, long-horizon embodied reasoning requires embodied spatio-temporal memory that connects visual evidence with motion, location, and time~\citep{anwar2025remembr,gorlo2026describe,hu2025threedllmmem,xie2024embodiedrag}. As agentic multimodal Retrieval-augmented Generation (RAG) has been recognized as an effective approach to retrieving and integrating multimodal evidence~\citep{yu2024visrag,jeong2025videorag,faysse2024colpali,fan2024videoagent,wang2024videoagent}, applying this paradigm to reasoning over embodied spatio-temporal memory is a promising direction. 

However, we identify three challenges in reasoning over long-horizon spatio-temporal memory:
\textbf{\textit{1)~Fragmented visual and motion context.}} Memory representations must preserve visual scene details while linking observations to motion context, including turns and elapsed time. Furthermore, evidence retrieval from structured embodied experience presents the second challenge:
\textbf{\textit{2)~Misalignment between semantic relevance and spatio-temporal constraints.}} Retrieval must distinguish semantically similar observations collected at different locations or times to satisfy the query's spatio-temporal constraints. Finally, using the retrieved evidence presents the third challenge:
\textbf{\textit{3)~Insufficient task awareness in context shaping.}} Tasks such as spatial comparison and elapsed-time estimation require different ways of organizing evidence in context. Task requirements must also guide further retrieval and context updates as reasoning proceeds. These challenges motivate the design of our framework for reasoning over embodied experience.

In this paper, we propose \mbox{\textbf{VimRobo}}, a framework that reconstructs task-relevant scene relations in context for reasoning over long-horizon embodied experience, as shown in Figure~\ref{fig:framework}.
\textbf{\textit{(i) Visual and Motion Memory Construction}} pairs visual evidence indexed by content, position, and time with compact motion descriptions of route segments, turns, and elapsed times. This memory preserves fine-grained scene details and connects observations with motion context, supporting reasoning across space and time.
\textbf{\textit{(ii) Multi-Perspective Memory Retrieval}} combines Spatio-Temporal Retrieval and Multi-Perspective Reranking to retrieve evidence and rerank semantic candidates against task-specific spatial or temporal anchors. This improves evidence localization by distinguishing semantically similar observations collected at different locations or times.
\textbf{\textit{(iii) In-Context Scene Relation Reconstruction}} uses task-specific skills to guide evidence selection, visual verification, and context updates. The agent retrieves necessary evidence according to task requirements and retains the relevant observations with spatio-temporal cues in context, organizing them into an evidence graph. This keeps supporting evidence available across reasoning steps, enabling task-specific reconstruction of scene relations as new evidence is retrieved, verified, and added to context.

Moreover, for training and evaluation, we propose a Q\&A synthesis pipeline using an instance graph to link observations across segments. Spatial, temporal, and event relations guide question generation, providing supervision for reasoning across objects. We construct RoboMemBench as a manually verified complementary diagnostic set to assess spatio-temporal evidence integration across object instances.

Our major contributions are as follows:
\begin{itemize}
  \item We conduct a pilot study of embodied reasoning, identifying how retrieval perspectives and memory representations influence visual evidence localization and answer accuracy.
  \item We construct visual and motion memory and design multi-perspective retrieval to distinguish semantically similar observations using spatial or temporal task anchors.
  \item We introduce an in-context scene relation reconstruction paradigm, with task-specific embodied skills guiding evidence selection, visual verification, and context updates.
  \item We develop an embodied Q\&A synthesis pipeline for training and evaluation, and construct RoboMemBench for challenging cross-instance relational reasoning.
\end{itemize}

\section{Pilot Study: Task and Motivation}
\label{sec:pilot}

In this section, we investigate critical challenges in embodied experience reasoning to motivate VimRobo in Section~\ref{sec:vimrobo}.

\subsection{Preliminary}

\textbf{Task Definition.}
We study embodied experience reasoning, where an agent answers a query $q$ about a robot's recorded visual observations and motion $\mathcal{D}$. The answer $\hat{y}=f(q,\mathcal{D})$ is a factual description, spatial coordinates, or a time value.

\noindent\textbf{Overview of Framework.}
We distinguish two paradigms by how memory is formed from experience:\\
\textit{(i)~\mbox{Offline Memory Modeling}.}
Existing methods~\citep{gu2024conceptgraphs,gorlo2026describe} typically construct static memory representations offline from Experience $\mathcal{D}$. The stored objects and relations remain fixed during task-specific processing and reasoning.\\
\textit{(ii)~\mbox{Online In-Context Memory Reconstruction}.}
We implement this online paradigm by iteratively retrieving evidence and reconstructing task-relevant scene relations in context. The agent generates an action $a_k$ based on context $\mathcal{H}_k$, obtains observations $\mathcal{O}_k$, and updates scene relations with relevant evidence. The evolving context preserves supporting observations across reasoning steps and guides further retrieval and visual verification according to task requirements.
\par

\subsection{\texorpdfstring{Impact of Memory Representation in Reasoning}{Impact of Memory Representation in Reasoning}}
\label{sec:pilot_context}

We investigate how motion and visual memory representations affect reasoning accuracy and computational cost.

\textbf{Experimental Settings.}
We compare three motion and three visual evidence representations using Qwen3.5-9B on timing and instance-extraction tasks. For motion, we vary the trajectory representation in context while holding trajectories and observations fixed. For visual evidence representations, we compare retained observations with text captions generated from them and graphs constructed from those captions. 

\textbf{Observations.}
As shown in Table~\ref{tab:pilot_compression}, for motion memory, converting numerical sequences into natural language improves accuracy with fewer tokens than coordinates and visual maps. This makes motion events and temporal relations explicit in prompts and aligns with the model's language reasoning capabilities. For visual evidence representations, pure vision achieves higher accuracy than captions and graphs at lower computational cost. The visual evidence in its raw modality supports query-conditioned perception without the semantic compression caused by captions or graphs.

\begin{table}[!t]
    \centering
    \caption{Memory representations for motion in context and visual evidence in storage. Left: we encode identical motion trajectories as numerical coordinates, visual maps, or natural language (NL). Right: we evaluate instance extraction on visual input (V), kept or converted to text captions (T) or caption-based graphs (G).}
    \label{tab:pilot_compression}
    \begingroup
    \fontsize{9}{11}\selectfont
    \setlength{\tabcolsep}{3pt}
    \renewcommand{\arraystretch}{1.15}
    \newcommand{\pilotmemoryheader}[1]{%
        \makecell[l]{\textbf{Memory}\\\textbf{Representation}} &
        \makecell[l]{\textbf{Information}\\\textbf{Transformation}} &
        \textbf{#1}\,$\downarrow$ &
        \textbf{Acc.}\,$\uparrow$ \\}
    \begin{minipage}[t]{0.505\linewidth}
        \begin{tabular*}{\linewidth}{@{}l@{\extracolsep{\fill}}lrr@{}}
            \toprule
            \pilotmemoryheader{Tokens}
            \midrule
            Coordinates & Direct Numbers & 12,046 & 17.3 \\
            Visual Map & Numbers $\to$ Map & 1,553 & 18.7 \\
            Motion Description & Numbers $\to$ NL & \textbf{1,221} & \textbf{23.3} \\
            \bottomrule
        \end{tabular*}
    \end{minipage}\hfill
    \begin{minipage}[t]{0.475\linewidth}
        \begin{tabular*}{\linewidth}{@{}l@{\extracolsep{\fill}}lrr@{}}
            \toprule
            \pilotmemoryheader{PFLOPs}
            \midrule
            Captions & V $\to$ T & 5.057 & 54.1 \\
            Graph & V $\to$ T $\to$ G & 7.639 & 45.9 \\
            Visual Memory & Retained V & \textbf{1.433} & \textbf{59.5} \\
            \bottomrule
        \end{tabular*}
    \end{minipage}
    \endgroup
\end{table}

\textbf{Insights.}
Matching representations to information characteristics supports efficient embodied reasoning, particularly when combining coarse-grained yet lengthy numerical motion records with fine-grained visual evidence.

\subsection{Impact of Retrieval Perspective in Localization}
\label{sec:pilot_retrieval}

\begin{wrapfigure}{r}{0.6\linewidth}
    \centering
    \vspace{-12pt}
    \includegraphics[width=\linewidth]{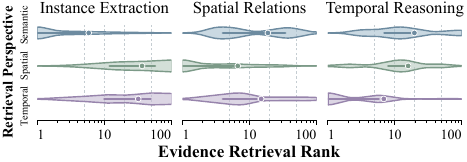}
    \vspace{-23pt}
    \caption{Comparison of retrieval perspectives for embodied evidence localization. We compare retrieval rank distributions (farther left is better) across three localization tasks, demonstrating that different retrieval perspectives favor specific task patterns.}
    \label{fig:retrieval_perspectives}
    \vspace{-8pt}
\end{wrapfigure}

We investigate how retrieval perspectives affect evidence localization in embodied spatio-temporal reasoning tasks.

\textbf{Experimental Settings.}
We compare semantic, spatial, and temporal retrieval on shared visual memory, with spatial and temporal anchors from queries, retrieved evidence, or trajectories. We evaluate first-hit ranks for instance extraction, spatial relations, and temporal reasoning.

\textbf{Observations.}
As shown in Figure~\ref{fig:retrieval_perspectives}, semantic retrieval concentrates early evidence hits in instance extraction. Spatial retrieval favors spatial relations, while temporal retrieval ranks relevant observations earlier for temporal reasoning. Each perspective provides a useful entry point for its matching task. Within these task categories, the distributions span a wide range, showing that retrieval effectiveness varies with the evidence required by each individual question.

\textbf{Insights.}
Different question patterns require distinct retrieval capabilities, motivating better integration of semantic, spatial, and temporal retrieval tools for complex tasks.

\par

\section{VimRobo}
\label{sec:vimrobo}

We introduce \textbf{VimRobo} for online scene relation reconstruction over embodied experience. We construct structured visual memory $\mathcal{C}=\{e_i\}_{i=1}^{N}$ and motion memory (\S~\ref{subsec:vimrobo_formulation}). Multi-Perspective Memory Retrieval (\S~\ref{subsec:vimrobo_skills}) combines search with anchor-based semantic reranking. The agent iteratively organizes and visually verifies evidence with task-specific skills via In-Context Scene Relation Reconstruction (\S~\ref{subsec:vimrobo_graph}).

\subsection{Visual and Motion Memory Construction}
\label{subsec:vimrobo_formulation}

We use visual memory for retrieval and motion memory for context shaping, following the \textit{1$^{st}$~insight} in Section \ref{sec:pilot_context}.

\textbf{Construction of Long-term Visual Memory.} We retain visual memory to preserve fine-grained details and avoid the cost of caption generation. We sample video segments from memory and encode each with a visual embedding model. Each memory entity pairs a clip with its retrieval indices:
\begin{compactmath}
\begin{equation}
    e_i \triangleq
    (\mathbf{I}_i,\mathbf{v}_i,p_i,t_i),
\end{equation}
\end{compactmath}
where $\mathbf{I}_i$ is the visual segment, $\mathbf{v}_i$ its embedding, and $p_i$ and $t_i$ its mean coordinate and timestamp. These indices support semantic, spatial, and temporal retrieval of visual evidence.

\textbf{Context Shaping of Motion Memory.} We convert the global trajectory $\Gamma=\{(p_\ell,t_\ell)\}_{\ell=1}^{L}$ into compact motion context using Ramer--Douglas--Peucker (RDP) simplification~\citep{douglas1973algorithms}. For endpoints $p_a,p_b$, the farthest interior point $p_{j^\star}$ has perpendicular distance $\delta_{ab}$ from the line joining them:
\begin{compactmath}
\begin{equation}
    \bar{\Gamma}_{a:b}=
    \begin{cases}
        \{(p_a,t_a),(p_b,t_b)\}, & \delta_{ab}\leq\varepsilon_{\mathrm{RDP}},\\
        \bar{\Gamma}_{a:j^\star}\cup\bar{\Gamma}_{j^\star:b}, & \delta_{ab}>\varepsilon_{\mathrm{RDP}},
    \end{cases}
\end{equation}
\end{compactmath}
where $\varepsilon_{\mathrm{RDP}}$ bounds geometric deviation, and $\delta_{ab}=0$ for adjacent endpoints. The resulting $\bar{\Gamma}=\bar{\Gamma}_{1:L}$ retains keypoints $(\bar{p}_m,\overline{t}_m)$ in temporal order. We calculate Euclidean displacements and elapsed times between consecutive keypoints:
\begin{compactmath}
\begin{equation}
    \Delta d_m=\|\bar{p}_{m+1}-\bar{p}_m\|_2,\qquad
    \Delta t_m=\overline{t}_{m+1}-\overline{t}_m.
\end{equation}
\end{compactmath}

We retain turns whose angles $\varphi_m$ and both adjacent segment lengths pass the thresholds, keeping the larger angle when merging nearby turns. Short descriptions of segments and turns form textual motion memory $\mathcal{M}_{\tau}$, e.g., ``move about $\Delta d_m$ meters, then turn left (if $\varphi_m>0$).'' This natural language representation helps the model understand motion better than numerical coordinate sequences or visual maps.

\subsection{Multi-Perspective Memory Retrieval}
\label{subsec:vimrobo_skills}

\begingroup
\let\retrievalcompactmath\compactmath
\renewcommand{\compactmath}{%
    \retrievalcompactmath
    \setlength{\abovedisplayskip}{4.5pt}%
    \setlength{\belowdisplayskip}{4.5pt}%
    \setlength{\abovedisplayshortskip}{0pt}%
    \setlength{\belowdisplayshortskip}{2pt}%
}

We follow the $\textit{2$^{nd}$~insight}$ in Section~\ref{sec:pilot_retrieval} to combine semantic, spatial, and temporal retrieval to locate evidence. We rerank semantic candidates using specific positions or timestamps as anchors while preserving target relevance. This design focuses retrieval on regions of interest in memory for more efficient localization with less unguided exploration.

\textbf{Structured Evidence Retrieval.} The agent searches memory $\mathcal{C}$ using position, time, or semantic similarity.

\begin{itemize}
\item \textbf{\textit{Position and Time Retrieval:}} The agent searches the full memory corpus near a reference position or timestamp $x$:
\begin{compactmath}
\begin{equation}
    \mathcal{R}_{\kappa}(x)
    =\operatorname{Top}_{K,\,e_i\in\mathcal{C}}
    [-d_{\kappa}(x_i,x)].
\end{equation}
\end{compactmath}
$\operatorname{Top}_{K}$ selects the $K$ highest-scoring entities. Here, $\kappa\in\{\mathrm{pos},\mathrm{time}\}$, with $d_{\mathrm{pos}}(p_i,p)=\|p_i-p\|_2$ on planar coordinates and $d_{\mathrm{time}}(t_i,t)=|t_i-t|$. These observations support spatial comparisons and temporal boundary verification.

\item \textbf{\textit{Semantic Anchoring-based Retrieval:}} This is a two-stage retrieval process: semantic retrieval followed by anchor-based reranking. The reranker acts like a fence for semantic search, using spatial or temporal anchors in the physical world to distinguish semantically similar candidates across locations or repeated visits. We encode query $z$ as text embedding $\mathbf{q}_{z}$ to retrieve the top $K$ semantic candidates:
\begin{compactmath}
\begin{equation}
    \mathcal{R}^{\mathrm{sem}}_{z}
    =\operatorname{Top}_{K,\,e_i\in\mathcal{C}}
    \operatorname{sim}(\mathbf{q}_{z},\mathbf{v}_{i}).
\end{equation}
\end{compactmath}
Then, the reranker sorts candidates by proximity to anchor $x_{\mathrm{anc}}$ from the query, retrieved evidence, or trajectory context, retaining only the $K_{\mathrm{r}}$ closest candidates, where $K_{\mathrm{r}}\leq K$:
\begin{compactmath}
\begin{equation}
    \mathcal{R}^{\mathrm{rerank}}_{z}
    =\operatorname{Top}_{K_{\mathrm{r}},\,e_i\in\mathcal{R}^{\mathrm{sem}}_{z}}
    [-d_{\kappa}(x_i,x_{\mathrm{anc}})].
\end{equation}
\end{compactmath}
For example, to find a chair near an entrance, the agent retrieves clips showing chairs, then reranks them by distance to the entrance. For a door seen around a turn, it reranks relevant clips by proximity to the turn time in motion memory. The agent visually verifies these views and retrieves more evidence when spatial or temporal relations remain unclear.
\end{itemize}

\textbf{Spatio-Temporal Contextualization.} At step $k$, retrieval returns memory entities $\mathcal{R}_k$. We sample $n_i$ visual frames from each segment $\mathbf{I}_i$ to form the stamped observations:
\begin{compactmath}
\begin{equation}
    \mathcal{O}_k
    =\bigcup_{e_i\in\mathcal{R}_k}
    \bigl\{(f_{ij},p_{ij},t_{ij})\bigr\}_{j=1}^{n_i},
\end{equation}
\end{compactmath}
where $f_{ij}$ is the $j$-th sampled frame, with position $p_{ij}$ and timestamp $t_{ij}$ shown as \texttt{[x,y,z]} and \texttt{HH:MM:SS}. We retain selected visual frames linked to evidence nodes mentioned in Section~\ref{subsec:vimrobo_graph} for perception and visual verification.
\par\endgroup

\subsection{In-Context Scene Relation Reconstruction}
\label{subsec:vimrobo_graph}

\begin{wrapfigure}[27]{r}{0.5\linewidth}
    \vspace{-10pt}
    \hrule height 0.8pt
    \vspace{3pt}
    \captionsetup{skip=0pt}
    \captionof{algorithm}{Inference Pipeline of VimRobo}
    \label{alg:vimrobo_rollout}
    \vspace{0pt}
    \hrule height 0.4pt
    \vspace{2pt}
    {\footnotesize\raggedright\linespread{1.05}\selectfont
    \begin{algorithmic}[1]
    \REQUIRE
    \begin{tabular}[t]{@{}l@{}}
    Task $q$, Policy $\pi_{\theta}$, Skills $\mathcal{K}_{\mathrm{all}}$, Environment $\mathcal{E}$\\
    Instructions $I_{\mathrm{init}},I_{\mathrm{evolve}},I_{\mathrm{perc}}$.
    \end{tabular}
    \ENSURE Structured answer $\hat{y}$ and evidence graph $\mathcal{G}$.
    \STATE \textcolor{gray}{// 1. Global trajectory context (Section \ref{subsec:vimrobo_formulation})}
    \STATE $\mathcal{M}_{\tau}\gets\mathcal{E}.\text{GlobalTrajectory}$
    \STATE \textcolor{gray}{// 2. Embodied Task-specific Initialization (Section \ref{subsec:vimrobo_graph})}
    \STATE $a^{\mathrm{init}}\sim\pi_\theta(\cdot\mid q,I_{\mathrm{init}},\mathcal{K}_{\mathrm{all}},\mathcal{M}_{\tau})$; $(\mathcal{G}_0, \mathcal{H}_1, \mathcal{K}_q)\gets a^{\mathrm{init}}$
    \WHILE{$k\leq k_{\max}$}
        \STATE \textcolor{gray}{// 3. In-Context Scene Relation Reconstruction (\S~\ref{subsec:vimrobo_graph})}
        \STATE $a_{k}\sim\pi_{\theta}(\cdot\mid\mathcal{H}_{k})$
        \IF{$a_{k}=a^{\mathrm{ret}}$}
            \STATE Evidence retrieval: $\mathcal{O}_{k}\gets\mathcal{E}.\text{MultiPerspectiveRet}(a_{k}^{\mathrm{ret}})$
            \STATE Context shaping: $\mathcal{H}_{k+1}\gets\mathcal{H}_{k}\cup\mathcal{O}_{k}$
        \ELSIF{$a_{k}=a^{\mathrm{perc}}$}
            \STATE Scene Perception: $a_k^{\mathrm{perc}}\sim\pi_{\theta}(\cdot\mid\mathcal{H}_{k},I_{\mathrm{perc}})$
            \STATE Evidence evolve: $\mathcal{G}_k\gets\mathcal{G}_{k-1}\oplus v_k(a_k^{\mathrm{perc}})$
        \ELSIF{$a_{k}=a^{\mathrm{ans}}$}
            \STATE Terminal projection: $\mathcal{G}_k\gets\mathcal{G}_{k-1}\oplus v_{\mathrm{ans}}(a_k^{\mathrm{ans}})$
            \STATE \textbf{return} $(\hat{y},\mathcal{G}_{k})$
        \ENDIF
        \STATE Multimodal context shaping: $\mathcal{F}_k\gets\bigcup_{v_i\in\mathcal{G}_k}\mathcal{F}(v_i)$
        \STATE $k\gets k+1$
    \ENDWHILE
    \end{algorithmic}
    }
    \hrule height 0.8pt
\end{wrapfigure}

Inspired by context management for RAG~\citep{wang2026vimrag} and computer-use agents~\citep{zhu2026hymem}, we adopt a process in which the agent retrieves evidence for in-context scene relation reconstruction with task-specific skills. This supports more accurate and efficient retrieval at lower reasoning cost by keeping the context organized and limiting unnecessary growth.

\begingroup
\setlength{\abovedisplayskip}{5pt}
\setlength{\belowdisplayskip}{5pt}
\setlength{\abovedisplayshortskip}{0pt}
\setlength{\belowdisplayshortskip}{3pt}
\textbf{Context Initialization with Task-Specific Skills.} We initialize the context for task query $q$ and use $a^{\mathrm{init}}$ to select skills $\mathcal{K}_q$ from $\mathcal{K}_{\mathrm{all}}$, distilled from training rollouts:
\begin{equation}
\label{eq:vimrobo_initialization}
    a^{\mathrm{init}}
    \sim\pi_\theta(\cdot\mid q,I_{\mathrm{init}},\mathcal{K}_{\mathrm{all}},\mathcal{M}_{\tau}),
\end{equation}
where $I_{\mathrm{init}}$ is the initial instruction and $\mathcal{M}_{\tau}$ is the motion memory. This allows the agent to focus on suitable skills.

\textbf{Context Shaping for Scene Relation Reconstruction.} As outlined in Algorithm~\ref{alg:vimrobo_rollout}, the agent shapes its context to guide evidence retrieval and scene relation reconstruction:
\begin{equation}
\label{eq:vimrobo_context}
    \mathcal{H}_k=(q,I_{\mathrm{evolve}},\mathcal{G}_{k-1},\mathcal{K}_q,\mathcal{M}_{\tau},\mathcal{F}_{k-1}),
\end{equation}
where $I_{\mathrm{evolve}}$ is the reconstruction instruction, $\mathcal{G}_{k-1}$ is scene relation presented in graph, $\mathcal{M}_{\tau}$ is the motion memory, and $\mathcal{F}_{k-1}$ retained relevant visual evidence. After retrieval $a_k^{\mathrm{ret}}$, the context is updated with observations $\mathcal{O}_k$:
\begin{equation}
\label{eq:vimrobo_context_update}
    \mathcal{H}_{k+1}=\mathcal{H}_k\cup\mathcal{O}_k.
\end{equation}
The agent then performs perception $a_{k+1}^{\mathrm{perc}}$ to update the context and complete one round of scene relation reconstruction. Each perception or answer action creates an evidence node $v_k$ that joins the graph as $\mathcal{G}_k=\mathcal{G}_{k-1}\oplus v_k$, where $\oplus$ links $v_k$ to its supporting parent nodes.

\textbf{Visual Evidence Verification via Scene Relation Reconstruction.} During perception, the model filters target-relevant frames through a binary judgment, then allocates vision tokens budgets to more important visual evidence:
\begin{equation}
\label{eq:vimrobo_pixel_weight}
    \omega_{i,j}=\min\{1+\operatorname{deg}^{+}_{\mathcal{G}_k}(v_i),c\}/m_i,
\end{equation}
Here, $m_i=|\mathcal{F}(v_i)|$ and $\operatorname{deg}^{+}$ counts direct references.
We use $c=5$ within a fixed pixel budget. Through context
shaping and visual verification, we retain key task evidence.

\par\endgroup

\section{Data Synthesis and Policy Optimization}
\label{sec:benchmark}

We synthesize Q\&A pairs with a two-stage pipeline (\S~\ref{subsec:robomem_synthesis}), then use them as seeds to implement VimRobo's iterative reconstruction paradigm using policy optimization (\S~\ref{subsec:vimrobo_rl}).

\Needspace{32\baselineskip}
\subsection{RoboMem Question-Answer Synthesis}
\label{subsec:robomem_synthesis}

\begin{wrapfigure}{r}{0.50\linewidth}
    \vspace{-20pt}
    \centering
    \includegraphics[width=\linewidth]{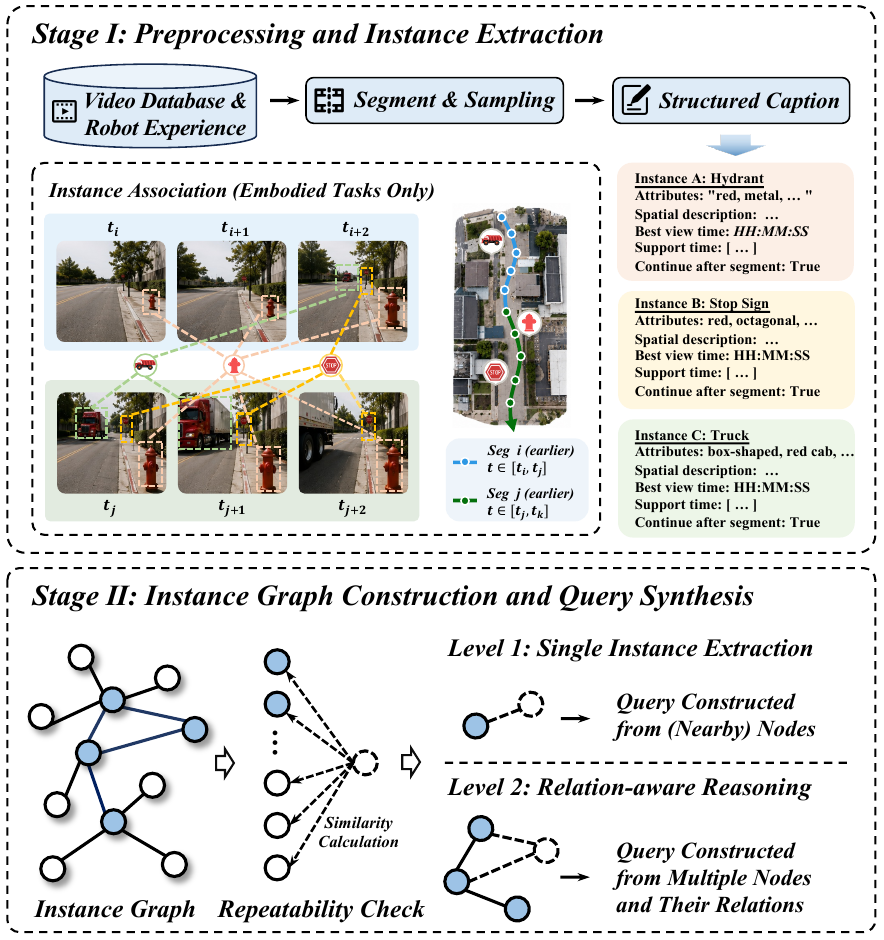}
    \vspace{-22pt}
    \caption{Data construction pipeline. We construct question-answer pairs in two stages. \textit{Stage I: Preprocessing and Instance Extraction.} Videos are split into overlapping segments to extract instances and events. Structured caption evidence is used to link the same instance across segments. \textit{Stage II: Graph Construction and Query Synthesis.} We maintain globally consistent instance IDs to construct an instance graph, whose edges represent spatio-temporal and event relations. Level 1 questions are built from one instance or its nearby nodes, while Level 2 questions require identifying and comparing multiple nodes and their spatial, temporal, and event relations.}
    \label{fig:data_construction}
    \vspace{-10pt}
\end{wrapfigure}

As shown in Figure~\ref{fig:data_construction}, we synthesize Q\&A pairs in two stages to help the model learn how to reconstruct spatial, temporal, and event relations in context from embodied experience.

\textbf{Stage I: Preprocessing and Instance Extraction.} At this stage, each video is divided into overlapping segments to preserve caption continuity, and structured captions extract visible instances and events. Each caption $c_\ell=\{x_a^\ell\}_a$ records spatial descriptions and observation times in segment $\ell$. We link instances $x_a^\ell \leftrightarrow x_b^{\ell+1}$ when their structured caption evidence indicates the same physical object, and assign linked instances a shared global identifier $\mathrm{id}_m$. The linked instances and event descriptions are stored in a global instance table, which supports query construction either within a physical environment or over a long video. This shared representation connects observations across time and provides linked evidence for question synthesis in next stage.

\textbf{Stage II: Instance Graph Construction and Query Synthesis.} We construct an instance graph $G_{\mathrm{inst}}=(V,E)$ from the global instance table produced in Stage I, where $V$ contains global instances and events, and $E$ includes caption-supported spatio-temporal and event relations. Before constructing queries, we retain as candidates only deduplicated instances with distinctive attributes through similarity-based matching and LLM-based semantic filtering. This process reduces ambiguous questions and incorrect answers caused by duplicate or confusable instances. \textit{Level 1} questions are generated from one candidate instance or its immediate neighbors and retrieve a direct fact. \textit{Level 2} questions are constructed by first expanding a candidate target into a connected local subgraph $G_q$. We then select a set of relation paths linking the target to other instances in $G_q$, and use the relations along these paths as constraints for question generation. Questions are retained only when the combined constraints uniquely identify the target, requiring the agent to compare instances or combine spatial, temporal, and event evidence across nodes. The pipeline supports descriptive questions and questions with structured answers, such as the best-view time or observation location. We manually verify all question-answer pairs sampled for RoboMemBench.

\subsection{Policy Learning for Embodied Reasoning}
\label{subsec:vimrobo_rl}

We perform supervised fine-tuning (SFT) and reinforcement learning (RL) to align reasoning and context shaping.

\begingroup
\setlength{\abovedisplayskip}{5pt}
\setlength{\belowdisplayskip}{5pt}
\setlength{\abovedisplayshortskip}{0pt}
\setlength{\belowdisplayshortskip}{3pt}
\textbf{Embodied Reasoning Trajectory Segmentation.} We train one policy on the question distribution synthesized in Section~\ref{subsec:robomem_synthesis}. We use the same segmentation for SFT warm-up and policy update in RL. We sample $G$ rollouts per query, each rollout $\tau_g=(\tau_g^{\mathrm{init}},\tau_g^{(1)},\ldots,\tau_g^{(n_g)})$ are segmented as:
\begin{equation}
\label{eq:vimrobo_segments}
\begin{alignedat}{2}
    &\tau_g^{\mathrm{init}} &&{}={} (q,I_{\mathrm{init}},\mathcal{K}_{\mathrm{all}},\mathcal{M}_{\tau};a_g^{\mathrm{init}}),\\
    &\tau_g^{(k)} &&{}={} (\mathcal{H}_{g,k};a^{\mathrm{ret}}_{g,k};\mathcal{O}_{g,k};a^{\mathrm{perc}}_{g,k}),\\
    &\tau_g^{(n_g)} &&{}={} (\mathcal{H}_{g,n_g};a^{\mathrm{ans}}_{g,n_g}),
\end{alignedat}
\end{equation}
where $\tau_g^{\mathrm{init}}$ selects task skills, $\mathcal{H}_{g,k}$ is the context for reconstruction in Section~\ref{subsec:vimrobo_graph}. For $k<n_g$, segments pair retrieval with perception, the terminal segment returns the answer.
\par\endgroup

\textbf{Spatio-Temporal Outcome Supervision.} We score the predicted answer $\hat{y}$ against reference $y$ by answer type:
\begin{compactmath}
\setlength{\abovedisplayskip}{4.5pt}
\setlength{\belowdisplayskip}{4.5pt}
\setlength{\abovedisplayshortskip}{0pt}
\setlength{\belowdisplayshortskip}{2pt}
\begin{equation}
\label{eq:vimrobo_accept}
    r_g
    =
    \begin{cases}
    \mathbb{I}\!\left[J_{\psi}(q,\hat{y},y)=1\right],
    & \text{Description},\\
    \mathbb{I}\!\left[\|\hat{p}-p\|_2
    \leq \varepsilon_{\mathrm{pos}}\right],
    & \text{Position},\\
    \mathbb{I}\!\left[|\hat{t}-t|
    \leq \varepsilon_{\mathrm{time}}\right],
    & \text{Time}.
    \end{cases}
\end{equation}
\end{compactmath}
Here, $J_{\psi}$ judges descriptive QA, while $(p,t)$ and $(\hat{p},\hat{t})$ denote reference and predicted positions and times. We set $\varepsilon_{\mathrm{pos}}=1.5$\,m and $\varepsilon_{\mathrm{time}}=10$\,s for training.

\textbf{Policy Optimization.} With the rewards above, we jointly optimize evidence retrieval, visual verification, and answer generation by applying GSPO~\citep{zheng2025group} to every retrieval--perception and answer segment in Eq.~(\ref{eq:vimrobo_segments}). Segments in rollout $g$ share the advantage $\widehat A_g=(r_g-\bar r)/(\sigma_r+\delta)$, where the mean $\bar r$ and standard deviation $\sigma_r$ are computed over the $G$ rollouts of the same query using Eq.~(\ref{eq:vimrobo_accept}) and $\delta$ is a stabilizer. For the $t$-th generated token $z_{g,k,t}$ of segment $k$ with history $h_{g,k,t}$ and $T_{g,k}$ generated tokens, the sequence ratio is $\rho_{g,k}=\exp\bigl[\tfrac{1}{T_{g,k}}\sum_{t}\log\tfrac{\pi_\theta(z_{g,k,t}\mid h_{g,k,t})}{\pi_{\mathrm{old}}(z_{g,k,t}\mid h_{g,k,t})}\bigr]$. We maximize $\mathbb{E}\bigl[\tfrac{1}{\sum_g n_g}\sum_{g}\sum_{k}\min\bigl(\rho_{g,k}\widehat A_g,\operatorname{clip}(\rho_{g,k},1-\varepsilon,1+\varepsilon)\widehat A_g\bigr)\bigr]$ over all segments, where $\varepsilon$ controls clipping. We use a KL coefficient of 0.001 and train with LoRA rank 32.

\vspace{-5pt}
\section{Experiments}
\label{sec:experiments}

\vspace{-5pt}

\subsection{Experimental Settings}
\label{sec:settings}

\textbf{Benchmarks and Metrics.}
We evaluate on NaVQA~\citep{anwar2025remembr} for short, medium, and long sequences and RoboMemBench for cross-instance reasoning on long sequences. We average three inference runs. For descriptive tasks, we use a binary model-based judge. For spatial and temporal tasks, we compute position and time errors against the reference answers and then report the percentage of predictions within each threshold.

\textbf{Baselines.}
(1)~Vanilla~RAG retrieves with the original question and answers directly from the returned evidence. (2)~\mbox{ReAct}~\citep{yao2022react} alternates reasoning and retrieval. (3)~MemAgent~\citep{yu2025memagent} processes observations sequentially and updates its memory till answer. (4)~Mem1~\citep{zhou2025mem1} alternates retrieval with updates to a compact memory state. (5)~ReMEmbR~\citep{anwar2025remembr} transforms the visual observation to captions for then perform retrieval and answer in textual modality. (6)~ConceptGraphs~\citep{gu2024conceptgraphs} organizes objects and their spatial relations in an open-vocabulary 3D scene graph. (7)~DAAAM~\citep{gorlo2026describe} builds a pre-constructed 4D scene graph for tool-based reasoning.

\textbf{Training and Inference.}
For embodied reasoning model, we use Qwen3.5-4B and Qwen3.5-9B~\citep{qwen3.5} for training and inference. For embodied experience representation, we use 3\,s clips sampled at 2\,fps across all methods.

\begin{table}[!t]
    \small
    \centering
    \caption{Experimental results on NaVQA. We compare VimRobo with embodied scene reconstruction and general agentic reasoning baselines. We report descriptive accuracy and evaluate spatial and temporal accuracy (\%, $\uparrow$) at two error thresholds for strict and relaxed evaluation, with spatial thresholds of 1\,m and 5\,m and temporal thresholds of 5\,s and 20\,s, and paired values presented as strict / relaxed.}
    \label{tab:results_navqa}
    \setlength{\tabcolsep}{4pt}
    \sisetup{group-digits=false, detect-weight=true}
    \adjustbox{max width=\linewidth}{
        \begin{tabular}{
            >{\normalsize}l|
            *{4}{>{\normalsize}S[table-format=2.1]}|
            *{4}{>{\normalsize}S[table-format=2.1]@{\,\textnormal{/}\,}>{\normalsize}S[table-format=2.1]}|
            *{4}{>{\normalsize}S[table-format=2.1]@{\,\textnormal{/}\,}>{\normalsize}S[table-format=2.1]}
        }
        \toprule
        \multirow{2}{*}{\small\textsc{\textbf{Method}}} &
        \multicolumn{4}{c|}{\small\textsc{\textbf{Descriptive Accuracy}}} &
        \multicolumn{8}{c|}{\small\textsc{\textbf{Spatial Accuracy}}} &
        \multicolumn{8}{c}{\small\textsc{\textbf{Temporal Accuracy}}} \\
        & \multicolumn{1}{c}{\small\textbf{Short}} & \multicolumn{1}{c}{\small\textbf{Medium}} &
        \multicolumn{1}{c}{\small\textbf{Long}} & \multicolumn{1}{c|}{\small\textbf{Overall}} &
        \multicolumn{2}{c}{\small\textbf{Short}} & \multicolumn{2}{c}{\small\textbf{Medium}} &
        \multicolumn{2}{c}{\small\textbf{Long}} & \multicolumn{2}{c|}{\small\textbf{Overall}} &
        \multicolumn{2}{c}{\small\textbf{Short}} & \multicolumn{2}{c}{\small\textbf{Medium}} &
        \multicolumn{2}{c}{\small\textbf{Long}} & \multicolumn{2}{c}{\small\textbf{Overall}} \\
        \midrule
        \multicolumn{21}{c}{\small\textit{Qwen3.5-4B}}\\
        \midrule
        Vanilla RAG & 52.4 & 63.2 & 60.0 & 58.0 & 10.5 & 10.5 & 0.0 & 5.0 & 3.1 & 6.3 & 4.2 & 7.0 & 33.3 & 44.4 & 16.7 & 41.7 & 22.2 & 38.9 & 23.1 & 41.0 \\
        ReAct~\citep{yao2022react} & 57.1 & 60.5 & 65.0 & 60.0 & 5.3 & 10.5 & 5.0 & 25.0 & 0.0 & 12.5 & 2.8 & 15.5 & 22.2 & 33.3 & 8.3 & 33.3 & 22.2 & 27.8 & 17.9 & 30.8 \\
        MemAgent~\citep{yu2025memagent} & 19.0 & 7.9 & 5.0 & 12.0 & 0.0 & 5.3 & 0.0 & 0.0 & 0.0 & 0.0 & 0.0 & 1.4 & 11.1 & 33.3 & 8.3 & 8.3 & 0.0 & 11.1 & 5.1 & 15.4 \\
        Mem1~\citep{zhou2025mem1} & 40.5 & 42.1 & 45.0 & 42.0 & 5.3 & 21.1 & 10.0 & 15.0 & 9.4 & 15.6 & 8.5 & 16.9 & 22.2 & 22.2 & 16.7 & 33.3 & 16.7 & 27.8 & 17.9 & 28.2 \\
        \midrule
        ReMEmbR~\citep{anwar2025remembr} & 33.3 & 52.6 & 60.0 & 46.0 & 5.3 & 15.8 & 10.0 & 20.0 & 15.6 & 31.3 & 11.3 & 23.9 & 33.3 & 55.6 & 33.3 & 50.0 & 33.3 & 50.0 & 33.3 & 51.3 \\
        ConceptGraphs~\citep{gu2024conceptgraphs} & 40.5 & 44.7 & 65.0 & 47.0 & 10.5 & 26.3 & 0.0 & 10.0 & 3.1 & 18.8 & 4.2 & 18.3 & \multicolumn{2}{c}{--} & \multicolumn{2}{c}{--} & \multicolumn{2}{c}{--} & \multicolumn{2}{c}{--} \\
        DAAAM~\citep{gorlo2026describe} & 40.5 & 39.5 & 50.0 & 42.0 & 15.8 & 47.4 & 0.0 & 20.0 & 12.5 & 25.0 & 9.9 & 29.6 & 44.4 & 55.6 & 41.7 & 50.0 & 22.2 & 33.3 & 33.3 & 43.6 \\
        \midrule
        \textbf{VimRobo~(Ours)} & 66.7 & 65.8 & 75.0 & \bfseries 68.0 & 15.8 & 57.9 & 15.0 & 50.0 & 15.6 & 43.8 & \bfseries 15.5 & \bfseries 49.3 & 55.6 & 66.7 & 50.0 & 66.7 & 38.9 & 55.6 & \bfseries 46.2 & \bfseries 61.5 \\
        \midrule
        \multicolumn{21}{c}{\small\textit{Qwen3.5-9B}}\\
        \midrule
        Vanilla RAG & 57.1 & 68.4 & 50.0 & 60.0 & 10.5 & 15.8 & 0.0 & 10.0 & 6.3 & 9.4 & 5.6 & 11.3 & 55.6 & 66.7 & 25.0 & 58.3 & 27.8 & 55.6 & 33.3 & 59.0 \\
        ReAct~\citep{yao2022react} & 66.7 & 63.2 & 75.0 & 67.0 & 5.3 & 10.5 & 10.0 & 40.0 & 0.0 & 18.8 & 4.2 & 22.5 & 33.3 & 44.4 & 25.0 & 41.7 & 22.2 & 33.3 & 25.6 & 38.5 \\
        MemAgent~\citep{yu2025memagent} & 23.8 & 10.5 & 5.0 & 15.0 & 5.3 & 5.3 & 5.0 & 5.0 & 0.0 & 0.0 & 2.8 & 2.8 & 22.2 & 33.3 & 8.3 & 16.7 & 5.6 & 11.1 & 10.3 & 17.9 \\
        Mem1~\citep{zhou2025mem1} & 42.9 & 50.0 & 55.0 & 48.0 & 5.3 & 21.1 & 5.0 & 25.0 & 6.3 & 15.6 & 5.6 & 19.7 & 33.3 & 44.4 & 25.0 & 33.3 & 16.7 & 38.9 & 23.1 & 38.5 \\
        \midrule
        ReMEmbR~\citep{anwar2025remembr} & 35.7 & 55.3 & 65.0 & 49.0 & 10.5 & 47.4 & 15.0 & 35.0 & 18.8 & 43.8 & 15.5 & 42.3 & 44.4 & 66.7 & 41.7 & 58.3 & 38.9 & 55.6 & 41.0 & 59.0 \\
        ConceptGraphs~\citep{gu2024conceptgraphs} & 45.2 & 47.4 & 65.0 & 50.0 & 10.5 & 31.6 & 0.0 & 25.0 & 12.5 & 31.3 & 8.5 & 29.6 & \multicolumn{2}{c}{--} & \multicolumn{2}{c}{--} & \multicolumn{2}{c}{--} & \multicolumn{2}{c}{--} \\
        DAAAM~\citep{gorlo2026describe} & 45.2 & 50.0 & 50.0 & 48.0 & 15.8 & 52.6 & 5.0 & 25.0 & 15.6 & 28.1 & 12.7 & 33.8 & 44.4 & 66.7 & 41.7 & 75.0 & 33.3 & 50.0 & 38.5 & 61.5 \\
        \midrule
        \textbf{VimRobo~(Ours)} & 69.0 & 71.1 & 80.0 & \bfseries 72.0 & 26.3 & 68.4 & 15.0 & 55.0 & 21.9 & 46.9 & \bfseries 21.1 & \bfseries 54.9 & 55.6 & 77.8 & 66.7 & 75.0 & 44.4 & 72.2 & \bfseries 53.8 & \bfseries 74.4 \\
        \bottomrule
        \end{tabular}
    }
\vspace{-10pt}
\end{table}

\Needspace{24\baselineskip}
\subsection{Results}
\label{sec:results}

\begin{wraptable}{r}{0.5\linewidth}
    \vspace{-8pt}
    \small
    \centering
    \caption{Experimental results on RoboMemBench. We compare VimRobo with embodied scene reconstruction and general agentic reasoning baselines on multi-hop cross-instance reasoning tasks.}
    \label{tab:robomembenchv2_results}
    \setlength{\tabcolsep}{4pt}
    \adjustbox{max width=\linewidth}{
        \begin{tabular}{>{\normalsize}l|
            S[table-format=2.1,detect-weight=true]
            S[table-format=2.1,detect-weight=true]@{\hspace{0.3em}/\hspace{0.3em}}S[table-format=2.1,table-number-alignment=left,detect-weight=true]|
            S[table-format=2.1,detect-weight=true]
            S[table-format=2.1,detect-weight=true]@{\hspace{0.3em}/\hspace{0.3em}}S[table-format=2.1,table-number-alignment=left,detect-weight=true]}
        \toprule
        \multirow{3}{*}{\centering\small\textsc{\textbf{Method}}} &
        \multicolumn{3}{c|}{\small\textsc{\textbf{Qwen3.5-4B}}} &
        \multicolumn{3}{c}{\small\textsc{\textbf{Qwen3.5-9B}}} \\
        \cmidrule(lr){2-4}\cmidrule(l){5-7}
        & \multicolumn{1}{c}{\small\textbf{Desc.}} &
        \multicolumn{2}{c|}{\small\textbf{Spat.\&Temp.}} &
        \multicolumn{1}{c}{\small\textbf{Desc.}} &
        \multicolumn{2}{c}{\small\textbf{Spat.\&Temp.}} \\
        & \multicolumn{1}{c}{\small\textbf{Binary}} &
        \multicolumn{2}{c|}{\small\textbf{Strict/Relaxed}} &
        \multicolumn{1}{c}{\small\textbf{Binary}} &
        \multicolumn{2}{c}{\small\textbf{Strict/Relaxed}} \\
        \midrule
        Vanilla RAG & 33.3 & 4.4 & 11.1 & 33.3 & 6.7 & 11.1 \\
        ReAct~\citep{yao2022react} & 20.0 & 6.7 & 13.3 & 40.0 & 11.1 & 13.3 \\
        MemAgent~\citep{yu2025memagent} & 6.7 & 0.0 & 0.0 & 26.7 & 2.2 & 2.2 \\
        Mem1~\citep{zhou2025mem1} & 20.0 & 11.1 & 17.8 & 26.7 & 11.1 & 24.4 \\
        \midrule
        ReMEmbR~\citep{anwar2025remembr} & 20.0 & 15.6 & 17.8 & 33.3 & 17.8 & 28.9 \\
        ConceptGraphs~\citep{gu2024conceptgraphs} & 0.0 & 2.2 & 4.4 & 13.3 & 6.7 & 13.3 \\
        DAAAM~\citep{gorlo2026describe} & 6.7 & 11.1 & 17.8 & 13.3 & 15.6 & 26.7 \\
        \midrule
        \textbf{VimRobo~(Ours)} & \bfseries 46.7 & \bfseries 20.0 & \bfseries 24.4 & \bfseries 60.0 & \bfseries 22.2 & \bfseries 46.7 \\
        \bottomrule
        \end{tabular}
    }
    \vspace{-8pt}
\end{wraptable}

\textbf{Main Results.}
In Table~\ref{tab:results_navqa}, VimRobo achieves the best overall performance on NaVQA at both model scales, with improvements across descriptive, spatial, and temporal questions. The temporal improvements persist on long trajectories, supporting reasoning over extended embodied experience. In Table~\ref{tab:robomembenchv2_results}, VimRobo achieves the highest descriptive and spatio-temporal accuracy on RoboMemBench at both scales. Improvements under strict and relaxed thresholds indicate more precise cross-instance reasoning. At the 9B scale, VimRobo improves relaxed spatio-temporal accuracy by 17.8 percentage points over ReMEmbR. These results support combining visual observations with spatial and temporal context to link evidence across instances. Table~\ref{tab:performance_traditional} reports the lowest mean spatial and temporal errors on valid answers at both scales, confirming more accurate position and time predictions alongside higher accuracy under both thresholds.

\begin{table}[!t]
    \centering
    \small
    \caption{Ablations. We organize the ablations around three mechanisms: evidence management (\S\,\ref{subsec:vimrobo_graph}) with in-context scene relation reconstruction (ISR) and visual evidence verification (VEV), task context (\S\,\ref{subsec:vimrobo_formulation}\,\&\,\ref{subsec:vimrobo_graph}) with motion memory (MOM) and task-specific skills (TSS), and memory retrieval (\S\,\ref{subsec:vimrobo_skills}) with spatio-temporal retrieval (STR) and multi-perspective reranking (MPR).}
    \label{tab:ablation_plus}
    \begingroup
    \footnotesize
    \setlength{\tabcolsep}{2.0pt}
    \renewcommand{\arraystretch}{1.0}
    \setlength{\aboverulesep}{0.3ex}
    \setlength{\belowrulesep}{0.45ex}
    \sisetup{
        group-digits=false,
        detect-weight=true,
        detect-family=true,
        table-format=2.1,
        table-number-alignment=center
    }
    \newcommand{\ablon}{\ding{51}}
    \newcommand{\abloff}{\textcolor{black!35}{\textendash}}
    \begin{tabular*}{\linewidth}{
        @{}c@{\hspace{0.25em}\extracolsep{\fill}}*{6}{w{c}{38pt}}
        @{\hspace{0.35em}}S
        @{\extracolsep{\fill}\hspace{0.3em}}S@{\extracolsep{0pt}\hspace{0.08em}/\hspace{0.08em}}S[table-number-alignment=left]
        @{\extracolsep{\fill}\hspace{0.3em}}S@{\extracolsep{0pt}\hspace{0.08em}/\hspace{0.08em}}S[table-number-alignment=left]@{}
    }
        \toprule
        &
        \multicolumn{2}{@{}c}{\footnotesize\makebox[0pt]{\textsc{\textbf{Evidence Management}}}} &
        \multicolumn{2}{c}{\footnotesize\makebox[0pt]{\textsc{\textbf{Task Context}}}} &
        \multicolumn{2}{c@{\hspace{0.35em}}}{\footnotesize\makebox[0pt]{\textsc{\textbf{Memory Retrieval}}}} &
        \multicolumn{1}{c}{\footnotesize\makebox[0pt]{\textsc{\textbf{Desc.}}}} &
        \multicolumn{2}{c}{\footnotesize\textsc{\textbf{Spatial}}} &
        \multicolumn{2}{c}{\footnotesize\textsc{\textbf{Temporal}}} \\
        \cmidrule(lr){2-3}\cmidrule(lr){4-5}\cmidrule(lr){6-7}
        \cmidrule(lr){8-8}\cmidrule(lr){9-10}\cmidrule(l){11-12}
        & \footnotesize\textbf{ISR} & \footnotesize\textbf{VEV} & \footnotesize\textbf{MOM} & \footnotesize\textbf{TSS} &
        \footnotesize\textbf{STR} & \footnotesize\textbf{MPR} &
        \multicolumn{1}{c}{\footnotesize\makebox[0pt]{\textbf{Binary}}} &
        \multicolumn{2}{c}{\footnotesize\textbf{@1\,/\,5\,m}} &
        \multicolumn{2}{c}{\footnotesize\textbf{@5\,/\,20\,s}} \\
        \midrule

        \multicolumn{12}{@{}l}{\scriptsize\itshape (a) Joint ablations} \\[0.5pt]
        \scriptsize (1) & \abloff & \abloff & \abloff & \abloff & \ablon & \abloff &
        55.7  & 6.3 & 30.2 & 15.3 & 45.8 \\
        \scriptsize (2) & \ablon & \ablon & \abloff & \ablon & \ablon & \abloff &
        65.2 & 12.5 & 46.9 & 28.8 & 59.3 \\

        \midrule
        \multicolumn{12}{@{}l}{\scriptsize\itshape (b) Experience retrieval ablations} \\[0.5pt]
        \scriptsize (3) & \ablon & \ablon & \ablon & \ablon & \abloff & \abloff &
        67.8 & 13.5 & 43.8 & 27.1 & 57.6 \\
        \scriptsize (4) & \ablon & \ablon & \ablon & \ablon & \abloff & \ablon &
        68.7 & 16.7 & 46.9 & 35.6 & 66.1 \\
        \scriptsize (5) & \ablon & \ablon & \ablon & \ablon & \ablon & \abloff &
        68.7 & 14.6 & 44.8 & 30.5 & 61.0 \\

        \midrule
        \multicolumn{12}{@{}l}{\scriptsize\itshape (c) Topology and task-state ablations} \\[0.5pt]
        \scriptsize (6) & \abloff & \abloff & \ablon & \ablon & \ablon & \ablon &
        58.3  & 7.3 & 34.4 & 20.3 & 50.8 \\
        \scriptsize (7) & \ablon & \abloff & \ablon & \ablon & \ablon & \ablon &
        67.8 & 11.5 & 46.9 & 33.9 & 59.3 \\
        \scriptsize (8) & \ablon & \ablon & \abloff & \ablon & \ablon & \ablon &
        69.6 & 14.6 & 49.0 & 30.5 & 62.7 \\
        \scriptsize (9) & \ablon & \ablon & \ablon & \abloff & \ablon & \ablon &
        66.1 & 12.5 & 47.9 & 37.3 & 59.3 \\
        \midrule
        & \ablon & \ablon & \ablon & \ablon & \ablon & \ablon &
        70.4 & 18.8 & 50.0 & 47.5 & 69.5 \\
        \bottomrule
    \end{tabular*}
    \endgroup
    \vspace{-10pt}
\end{table}

\textbf{Ablations.} Table~\ref{tab:ablation_plus} reports performance on both embodied memory benchmarks using the same backbone and training/inference budgets. With identical training data, VimRobo outperforms the reasoning--action paradigm~\citep{yao2022react} in Exp.~1. The graph-based RAG paradigm~\citep{wang2026vimrag} requires fine-tuning and prompt adjustment for embodied tasks. Exp.~2 is our closest implementation, fine-tuned on our data with our skills and tools. Its results support our training data's utility. These comparisons highlight \textbf{the need for our embodied in-context reconstruction to properly link relevant evidence.} Exps.~3 and~5 isolate STR, while Exps.~3 and~4 isolate MPR. VimRobo's gains over Exps.~4 and 5 demonstrate that \textbf{contextual search and semantic reranking jointly improve retrieval quality and efficiency.} Exps.~6 and~7 compare linear memory with ISR, keeping VEV disabled in both experiments. VimRobo's gains over Exps.~7~to~9 reflect VEV's priority for shared views over uniform allocation, MOM's motion context, and TSS's guidance, respectively. These gains show that \textbf{retrieval and visual evidence verification benefit from task-relevant evidence dependencies.}

\subsection{Analysis}
\label{sec:analysis}

\begin{wrapfigure}{r}{0.59\linewidth}
    \vspace{-10pt}
    \centering
    \includegraphics[width=\linewidth]{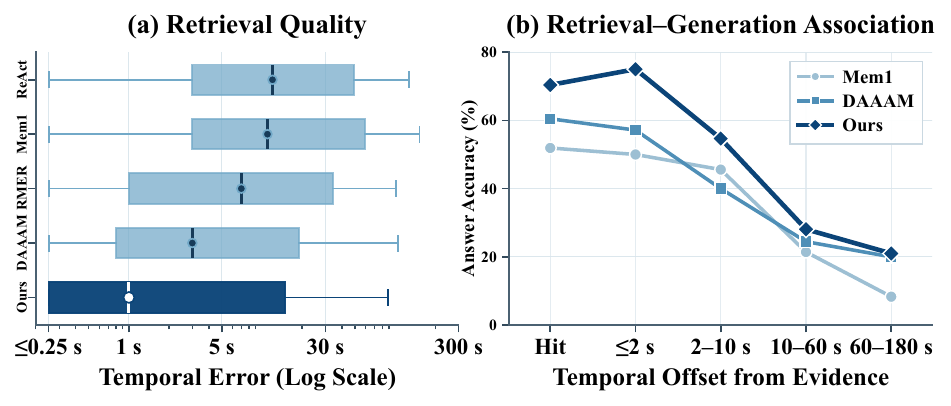}
    \vspace{-20pt}
    \caption{Retrieval quality and its relevance with generation. (a) Minimum temporal offsets to the reference frame. (b) Final-answer accuracy across retrieval outcomes and temporal-offset ranges.}
    \label{fig:retrieval_quality}
    \vspace{-8pt}
\end{wrapfigure}
\textbf{Spatio-temporal alignment improves retrieval and reasoning.} Spatial and temporal retrieval rank supporting observations earlier for their respective task categories in Figure~\ref{fig:retrieval_perspectives}. VimRobo retrieves evidence with smaller temporal offsets and achieves higher answer accuracy on retrieval hits and across offset ranges in Figure~\ref{fig:retrieval_quality}. The agent compares stamped observations to distinguish candidate locations and event order. Table~\ref{tab:ablation_plus} shows that retrieval by position and time complements reranking to improve temporal accuracy under the strict threshold, supporting accurate event localization.

\textbf{VimRobo predicts locations and times more precisely.} VimRobo achieves the lowest mean errors at both scales in Table~\ref{tab:performance_traditional}, consistent with its gains under strict and relaxed thresholds. Compared with DAAAM, it improves spatial and temporal precision at both scales, supporting more accurate predictions of locations and event times in embodied tasks.

\begin{table}[!t]
    \centering
    \begin{minipage}[t]{0.545\linewidth}
            \small
    \centering
    \begin{minipage}[t][76pt][t]{\linewidth}
    \captionof{table}{Spatial and temporal errors. We report mean errors over valid numerical answers. Spatial errors use planar Euclidean distance (m); temporal errors use absolute differences in timestamps or durations (s). Lower values indicate more accurate predictions.}
    \label{tab:performance_traditional}
    \end{minipage}
    \vspace{2pt}
    \setlength{\tabcolsep}{4pt}
    \adjustbox{max width=\linewidth}{
        \begin{tabular}{
            >{\normalsize}l|
            >{\normalsize}c
            >{\normalsize}c|
            >{\normalsize}c
            >{\normalsize}c
        }
        \toprule
        \multirow{3.5}{*}{\centering\small\textsc{\textbf{Method}}} &
        \multicolumn{2}{c|}{\small\textsc{\textbf{Qwen3.5-4B}}} &
        \multicolumn{2}{c}{\small\textsc{\textbf{Qwen3.5-9B}}} \\
        \cmidrule(lr){2-3}\cmidrule(l){4-5}
        & \small\textbf{Spatial} &
        \small\textbf{Temporal} &
        \small\textbf{Spatial} &
        \small\textbf{Temporal} \\
        & \small\textbf{Error (m)} &
        \small\textbf{Error (s)} &
        \small\textbf{Error (m)} &
        \small\textbf{Error (s)} \\
        \midrule
        Vanilla RAG & 83.5 & 192.3 & 74.1 & 179.9 \\
        ReAct~\citep{yao2022react} & 122.7 & 177.0 & 65.2 & 154.3 \\
        MemAgent~\citep{yu2025memagent} & 145.0 & 207.5 & 114.1 & 175.1 \\
        Mem1~\citep{zhou2025mem1} & 88.5 & 182.0 & 69.6 & 178.3 \\
        \midrule
        ReMEmbR~\citep{anwar2025remembr} & 49.5 & 167.0 & 37.1 & 147.3 \\
        ConceptGraphs~\citep{gu2024conceptgraphs} & 42.1 & -- & 34.0 & -- \\
        DAAAM~\citep{gorlo2026describe} & 33.2 & 156.3 & 29.4 & 124.9 \\
        \midrule
        \textbf{VimRobo~(Ours)} & \textbf{26.5} & \textbf{118.9} & \textbf{22.1} & \textbf{102.8} \\
        \bottomrule
        \end{tabular}
    }

    \end{minipage}\hfill
    \begin{minipage}[t]{0.435\linewidth}
            \centering
    \begin{minipage}[t][76pt][t]{\linewidth}
    \captionof{table}{Performance across training stages. End-to-end performance is measured by accuracy. Hit Rate ($\uparrow$) denotes the rate of retrieving supporting evidence. Success w/ Hit denotes the mean answer accuracy across answer types conditioned on a retrieval hit.}
    \label{tab:policy_optimization}
    \end{minipage}
    \vspace{4pt}
    \small
    \setlength{\tabcolsep}{4pt}
    \renewcommand{\arraystretch}{1.18}
    \begin{tabular*}{\linewidth}{@{}l@{\extracolsep{\fill}}cc@{}}
        \toprule
        & \textbf{SFT} & \textbf{+RL} \\
        \midrule
        \multicolumn{3}{@{}l}{\textit{End-to-end performance}} \\
        Descriptive & 65.2 & 70.4 \\
        Spatial & 12.5\,/\,46.9 & 18.8\,/\,50.0 \\
        Temporal & 32.2\,/\,55.9 & 47.5\,/\,69.5 \\
        \midrule
        \multicolumn{3}{@{}l}{\textit{Stage-wise performance}} \\
        Hit Rate & 70.3\,/\,77.8 & 75.2\,/\,80.7 \\
        Success w/ Hit & 30.0\,/\,45.6 & 33.7\,/\,51.9 \\
        \bottomrule
    \end{tabular*}

    \end{minipage}
    \vspace{-10pt}
\end{table}

\begin{wrapfigure}{r}{0.52\linewidth}
    \centering
    \includegraphics[width=\linewidth]{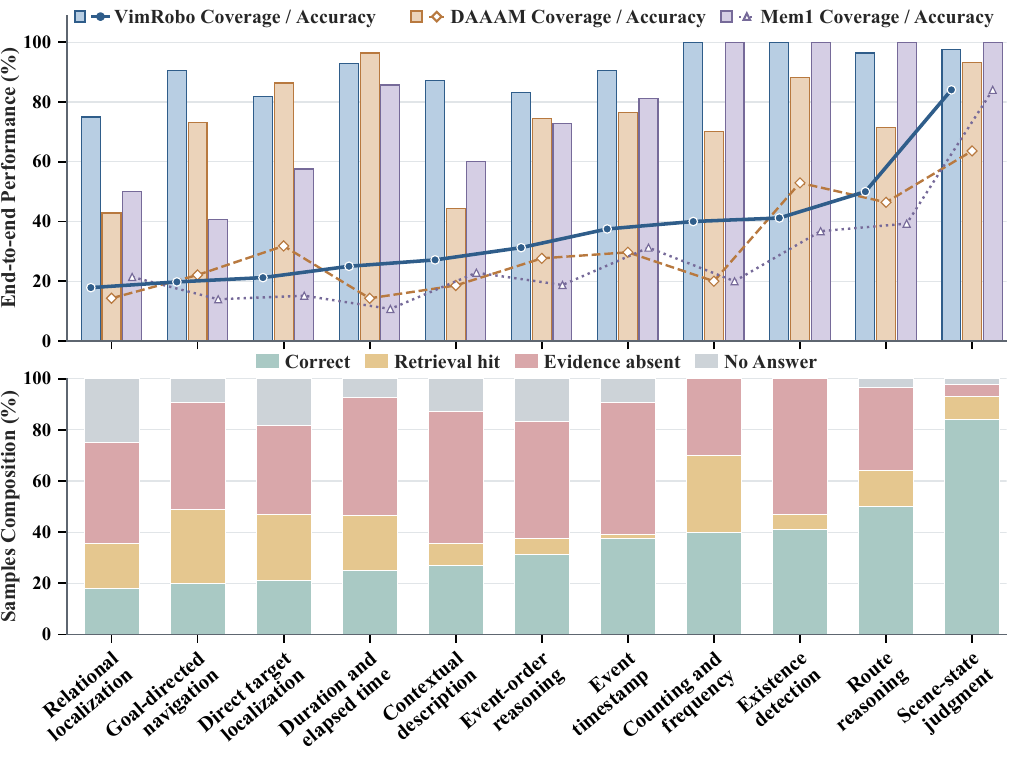}
    \vspace{-20pt}
    \caption{End-to-end performance decomposition and case distribution across task patterns. The upper panel reports end-to-end Coverage (bars) and Accuracy (lines), indicating higher task-level completion and accuracy. The lower panel shows VimRobo's distribution of Correct, Retrieval hit, Evidence absent, and No Answer cases, denoting successfully completed tasks, incorrect responses with retrieved supporting evidence, incorrect responses without hits in annotated support windows, and queries lacking a valid response within the allotted task budget, respectively.}
    \label{fig:task_pattern_performance}
\end{wrapfigure}

\textbf{Embodied context strengthens relational reasoning.} We improve contextual description, event ordering, and route reasoning in the upper panel of Figure~\ref{fig:task_pattern_performance}. These gains support linking observations across instances, positions, and times. Table~\ref{tab:ablation_plus} compares embodied and generic graph context management under matched training and pixel budgets. Global trajectory context brings larger gains to temporal accuracy than to descriptive answers. This pattern supports the role of motion history in connecting observations across time. The agent can use the recorded movement and elapsed time between observations to compare events at different locations and determine their order along the observed trajectory.

\Needspace{34\baselineskip}
\textbf{Policy optimization improves retrieval and evidence use.} Prior visual RAG uses reference-image retrieval rewards~\citep{wang2025vrag}, while graph-based optimization selects updates using answer-support paths and reference-aware masking~\citep{wang2026vimrag}. We use reference evidence to measure retrieval hits. In embodied reasoning, intermediate observations outside the annotated support window can also provide spatial anchors or temporal transitions needed to locate and verify the target. We therefore optimize all retrieval--perception and answer segments with GSPO on the synthesized task distribution, using the task outcome rewards in \S~\ref{subsec:vimrobo_rl}. Table~\ref{tab:policy_optimization} breaks down performance after SFT and subsequent RL. This improves accuracy across answer types, with higher retrieval hit rates and higher answer accuracy given a retrieval hit under both strict and relaxed spatial and temporal thresholds.

\textbf{Verification matters beyond retrieval hits.} The lower panel of Figure~\ref{fig:task_pattern_performance} shows that errors can occur even after supporting evidence is retrieved. The agent can revisit stamped views to compare candidates, check conflicting summaries, and verify relations before combining evidence. Parent references preserve dependencies among these observations as the graph grows. Tracking unresolved relations and search progress guides further retrieval when evidence is insufficient for the required spatial or temporal conclusion.

\begin{figure}[!t]
    \centering
    \includegraphics[width=1.0\linewidth]{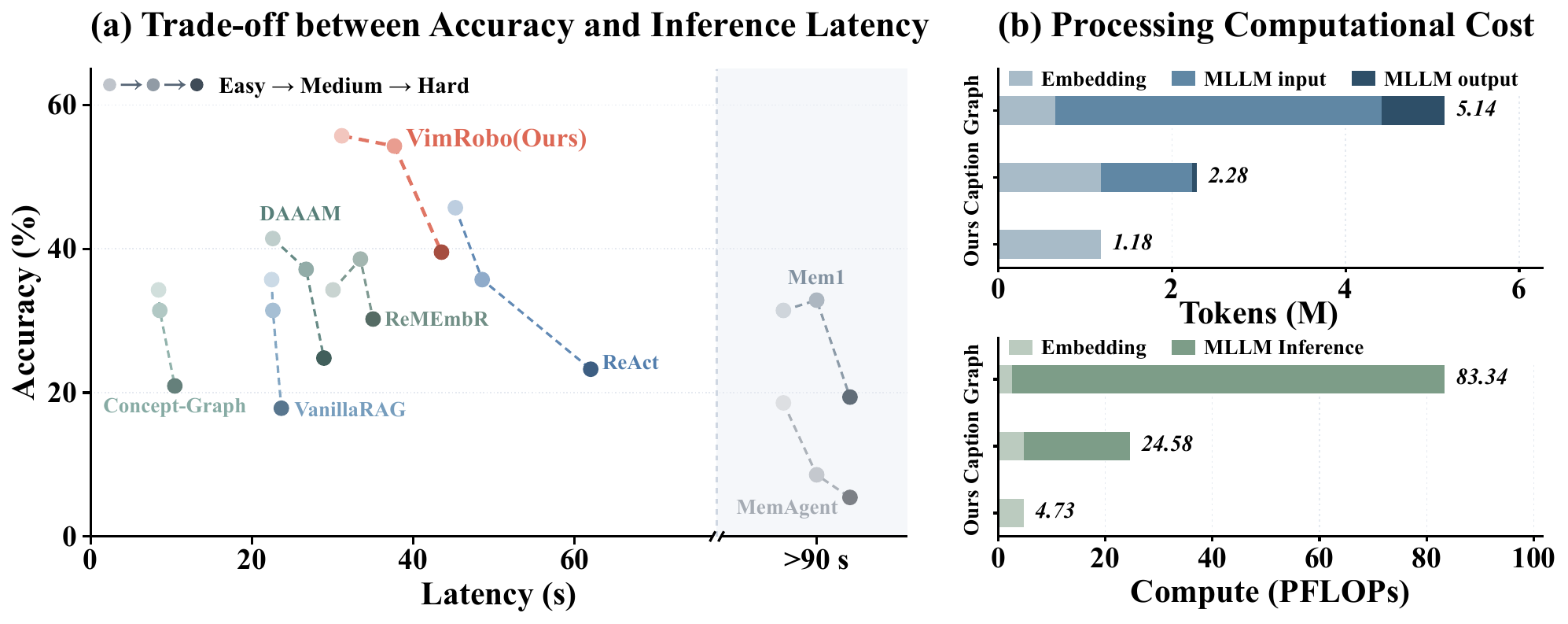}
    \vspace{-20pt}
    \caption{Inference and preprocessing efficiency. (a) Accuracy--latency trade-off, with darker markers indicating harder tasks. (b) Computational cost and token usage during memory construction.}
    \label{fig:inference_efficiency}
\end{figure}

\textbf{Generation latency and computational overhead.} VimRobo combines the highest accuracy with lower latency than ReAct and Mem1 across difficulty levels in Figure~\ref{fig:inference_efficiency}~(a). Embedding-based preprocessing also uses fewer tokens and less computation than caption-based and graph-based pipelines in Figure~\ref{fig:inference_efficiency}~(b). Deferring relation reconstruction to each task avoids corpus-wide language generation and directs visual verification toward task-relevant evidence.

\begin{figure}[!t]
    \centering
    \includegraphics[width=1\textwidth]{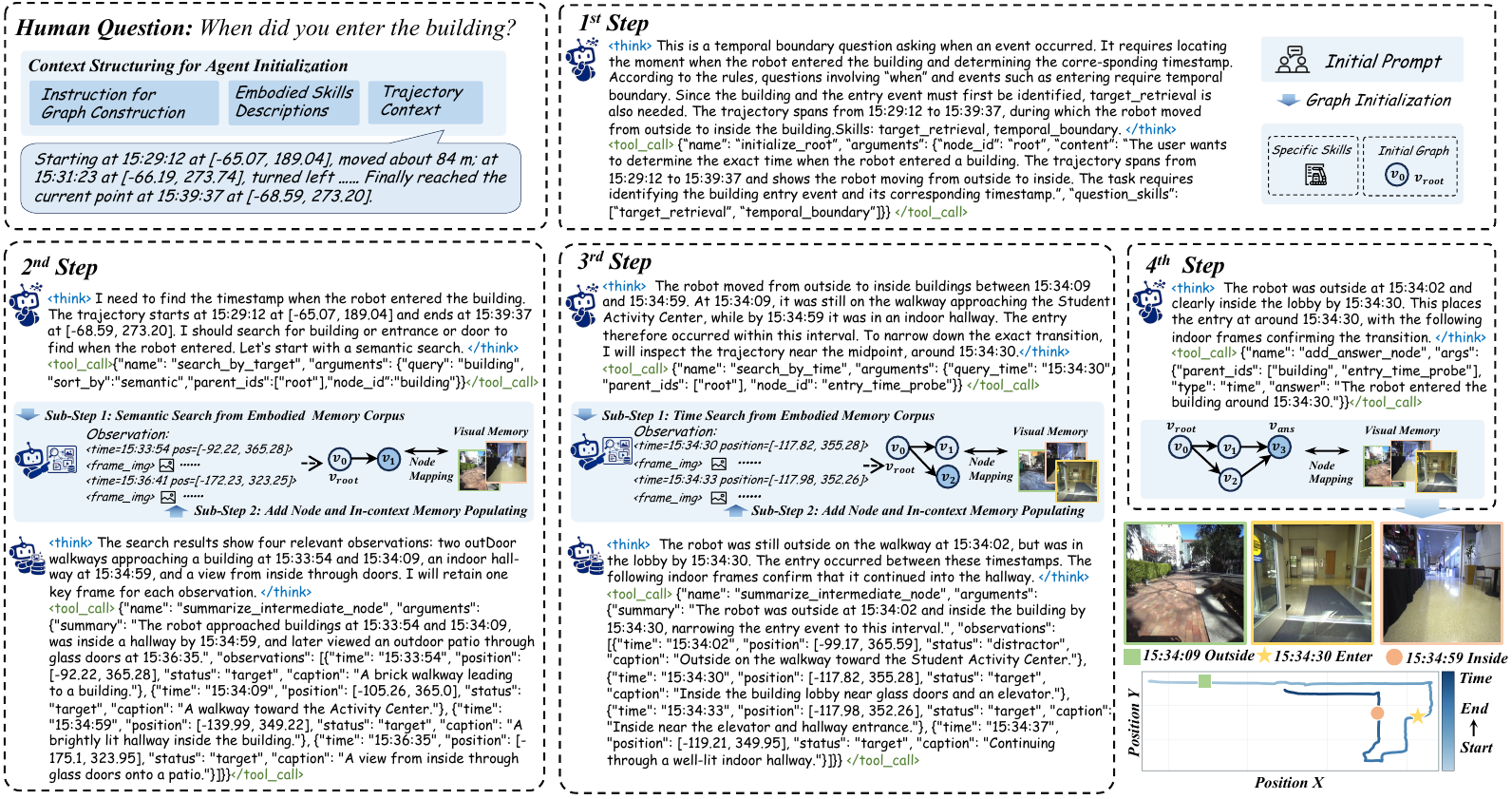}
    \vspace{-20pt}
    \caption{Case study. Semantic retrieval locates building views, temporal search verifies the entry time.}
    \label{fig:embodied_case_study}
    \vspace{-15pt}
\end{figure}

\textbf{Case study.} Figure~\ref{fig:embodied_case_study} illustrates how the agent combines semantic retrieval and temporal verification to localize a building-entry event. The agent verifies the outside-to-inside transition against the retrieved views and timestamps.

\section{Related Work}
\label{sec:relatedwork}

For in-context scene relation reconstruction, we review work on embodied memory and scene graphs, visual retrieval and video reasoning, and agent memory and policy learning.

\textbf{Embodied Memory and Scene Graphs.}
Embodied memories retrieve captioned observations through semantic lookup~\citep{anwar2025remembr,xie2024embodiedrag} or selectively attend to episodic visual features for long-horizon decisions~\citep{hu2025threedllmmem,fang2019scenememory}. Scene graphs organize object identities and spatial relations~\citep{gu2024conceptgraphs,werby2024hovsg,maggio2024clio,gorlo2026describe} to support memory-guided exploration and hierarchical planning across scene regions~\citep{saxena2025grapheqa,ginting2025mindpalace}. VimRobo reconstructs task-relevant scene relations in context from visual and motion memory.

\textbf{Visual Retrieval and Video Reasoning.}
Visual RAG retrieves images as answer evidence, while multimodal embeddings align text queries with images and videos~\citep{yu2024visrag,jeong2025videorag,faysse2024colpali,zhang2024gme,guo2025towards,li2026qwen3,wang2025vidorag,meng2025vlm2vec,gao2026scaling}. Video agents search query-relevant frames, events, or hierarchical segment summaries~\citep{fan2024videoagent,wang2024videoagent,ye2025temporalsearch,wang2025videotree}, while graph-structured memory preserves reasoning dependencies~\citep{wang2026vimrag}. VimRobo reranks semantic candidates against spatial or temporal task anchors to distinguish similar visual observations.

\textbf{Agent Memory and Policy Learning.}
Agents interleave reasoning and tool use~\citep{yao2022react} and retrieve compact gist summaries or linked memory records~\citep{lee2024readagent,gutierrez2024hipporag,gutierrez2025hipporag2,xu2025mem,kang2025memoryos,zhang2025survey}. Learned policies select search actions based on query needs~\citep{jin2025searchr1,asai2024self} and relevant observations to retain~\citep{zhou2025mem1,yu2025memagent,zou2026taskmem}. VimRobo uses task-specific skills and spatio-temporal outcome supervision to guide retrieval, visual verification, and context updates.

\section{Conclusion}
\label{sec:conclusion}

We proposed VimRobo to unify visual and motion memory, multi-perspective retrieval, and in-context scene relation reconstruction, with our Q\&A synthesis pipeline supporting its implementation. Experiments on various embodied memory benchmarks show gains in descriptive, spatial, and temporal reasoning, highlighting the benefits of evidence integration.

In our method, the main limitations lie in embedding quality and the base model's spatial reasoning. In the future, we will prioritize faithful memory construction, precise retrieval, and accurate interpretation of retrieved experience. Advances in the underlying model's perception and tool use are central to jointly strengthening these capabilities and improving overall system performance in embodied tasks.

\bibliographystyle{IEEEtran}
\bibliography{references}

@IEEEtranBSTCTL{IEEEexample:BSTcontrol,
  CTLuse_url = "no",
  CTLuse_forced_etal = "no",
  CTLmax_names_forced_etal = "10",
  CTLnames_show_etal = "3"
}

@inproceedings{wang2025vidorag,
  title={{ViDoRAG}: Visual document retrieval-augmented generation via dynamic iterative reasoning agents},
  author={Wang, Qiuchen and Ding, Ruixue and Chen, Zehui and Wu, Weiqi and Wang, Shihang and Xie, Pengjun and Zhao, Feng},
  booktitle={Conference on Empirical Methods in Natural Language Processing},
  year={2025},
  doi={10.18653/v1/2025.emnlp-main.464},
  url={https://aclanthology.org/2025.emnlp-main.464/}
}

@inproceedings{wang2026vimrag,
  title={Navigating Massive Visual Context in Retrieval-Augmented Generation via Multimodal Memory Graph},
  author={Qiuchen Wang and Shihang Wang and Yu Zeng and Qiang Zhang and Fanrui Zhang and Zhuoning Guo and Bosi Zhang and Wenxuan Huang and Lin Chen and Zehui Chen and Pengjun Xie and Ruixue Ding},
  booktitle={International Conference on Machine Learning},
  year={2026},
  url={https://proceedings.mlr.press/v306/wang26ip.html}
}

@inproceedings{wang2025vrag,
  title={{VRAG-RL}: Empower Vision-Perception-Based {RAG} for Visually Rich Information Understanding via Iterative Reasoning with Reinforcement Learning},
  author={Wang, Qiuchen and Ding, Ruixue and Zeng, Yu and Chen, Zehui and Chen, Lin and Wang, Shihang and Xie, Pengjun and Huang, Fei and Zhao, Feng},
  booktitle={Advances in Neural Information Processing Systems},
  year={2025},
  url={https://proceedings.neurips.cc/paper_files/paper/2025/hash/529b876b5d87f997e98ed925a3ce9f01-Abstract-Conference.html}
}

@inproceedings{guo2025towards,
  title={Towards Universal Video Retrieval: Generalizing Video Embedding via Synthesized Multimodal Pyramid Curriculum},
  author={Guo, Zhuoning and Li, Mingxin and Zhang, Yanzhao and Long, Dingkun and Xie, Pengjun and Chu, Xiaowen},
  booktitle={ACM International Conference on Multimedia},
  year={2026},
  url={https://arxiv.org/abs/2510.27571}
}

@inproceedings{zhang2024gme,
  title={Bridging Modalities: Improving Universal Multimodal Retrieval by Multimodal Large Language Models},
  author={Zhang, Xin and Zhang, Yanzhao and Xie, Wen and Li, Mingxin and Dai, Ziqi and Long, Dingkun and Xie, Pengjun and Zhang, Meishan and Li, Wenjie and Zhang, Min},
  booktitle={IEEE/CVF Computer Vision and Pattern Recognition Conference},
  year={2025},
  url={https://openaccess.thecvf.com/content/CVPR2025/html/Zhang_Bridging_Modalities_Improving_Universal_Multimodal_Retrieval_by_Multimodal_Large_Language_CVPR_2025_paper.html}
}

@article{li2026qwen3,
  title={{Qwen3-VL-Embedding} and {Qwen3-VL-Reranker}: A Unified Framework for State-of-the-Art Multimodal Retrieval and Ranking},
  author={Li, Mingxin and Zhang, Yanzhao and Long, Dingkun and Chen, Keqin and Song, Sibo and Bai, Shuai and Yang, Zhibo and Xie, Pengjun and Yang, An and Liu, Dayiheng and others},
  journal={arXiv preprint arXiv:2601.04720},
  year={2026}
}

@article{Qwen3-VL,
      title={{Qwen3-VL} Technical Report},
      author={Shuai Bai and Yuxuan Cai and Ruizhe Chen and Keqin Chen and Xionghui Chen and Zesen Cheng and Lianghao Deng and others},
	  journal={arXiv preprint arXiv:2511.21631},
      year={2025}
}

@inproceedings{fan2024videoagent,
  title={{VideoAgent}: A Memory-Augmented Multimodal Agent for Video Understanding},
  author={Fan, Yue and Ma, Xiaojian and Wu, Rujie and Du, Yuntao and Li, Jiaqi and Gao, Zhi and Li, Qing},
  booktitle={European Conference on Computer Vision},
  year={2024},
  url={https://www.ecva.net/papers/eccv_2024/papers_ECCV/html/3241_ECCV_2024_paper.php}
}

@inproceedings{zhou2025mem1,
  title={{MEM1}: Learning to Synergize Memory and Reasoning for Efficient Long-Horizon Agents},
  author={Zhou, Zijian and Qu, Ao and Wu, Zhaoxuan and Kim, Sunghwan and Prakash, Alok and Rus, Daniela and Low, Bryan Kian Hsiang and Liang, Paul},
  booktitle={International Conference on Learning Representations},
  year={2026},
  url={https://proceedings.iclr.cc/paper_files/paper/2026/hash/5fc8b3bdfbb9167b5144df5d3fae4616-Abstract-Conference.html}
}

@inproceedings{yu2025memagent,
  title={{MemAgent}: Reshaping Long-Context {LLM} with Multi-Conv {RL}-based Memory Agent},
  author={Yu, Hongli and Chen, Tinghong and Feng, Jiangtao and Chen, Jiangjie and Dai, Weinan and Yu, Qiying and Zhang, Ya-Qin and Ma, Wei-Ying and Liu, Jingjing and Wang, Mingxuan and Zhou, Hao},
  booktitle={International Conference on Learning Representations},
  year={2026},
  url={https://proceedings.iclr.cc/paper_files/paper/2026/hash/4264ee4376776907c0b87ed70b959585-Abstract-Conference.html}
}

@inproceedings{jeong2025videorag,
  title={{VideoRAG}: Retrieval-Augmented Generation over Video Corpus},
  author={Jeong, Soyeong and Kim, Kangsan and Baek, Jinheon and Hwang, Sung Ju},
  booktitle={Findings of the Association for Computational Linguistics: ACL 2025},
  year={2025},
  doi={10.18653/v1/2025.findings-acl.1096},
  url={https://aclanthology.org/2025.findings-acl.1096/}
}

@article{meng2025vlm2vec,
  title={{VLM2Vec-V2}: Advancing multimodal embedding for videos, images, and visual documents},
  author={Meng, Rui and Jiang, Ziyan and Liu, Ye and Su, Mingyi and Yang, Xinyi and Fu, Yuepeng and Qin, Can and Thirukovalluru, Raghuveer and Zhang, Xuan and Chen, Zeyuan and Xu, Ran and Xiong, Caiming and Zhou, Yingbo and Chen, Wenhu and Yavuz, Semih},
  journal={Transactions on Machine Learning Research},
  year={2026},
  url={https://openreview.net/forum?id=TpU38jbKIJ}
}

@inproceedings{yao2022react,
  title={{ReAct}: Synergizing Reasoning and Acting in Language Models},
  author={Yao, Shunyu and Zhao, Jeffrey and Yu, Dian and Du, Nan and Shafran, Izhak and Narasimhan, Karthik and Cao, Yuan},
  booktitle={International Conference on Learning Representations},
  year={2023},
  url={https://openreview.net/forum?id=WE_vluYUL-X}
}

@article{zheng2025group,
  title={Group sequence policy optimization},
  author={Zheng, Chujie and Liu, Shixuan and Li, Mingze and Chen, Xiong-Hui and Yu, Bowen and Gao, Chang and Dang, Kai and Liu, Yuqiong and Men, Rui and Yang, An and others},
  journal={arXiv preprint arXiv:2507.18071},
  year={2025}
}

@inproceedings{xu2025mem,
  title={{A-Mem}: Agentic Memory for {LLM} Agents},
  author={Xu, Wujiang and Liang, Zujie and Mei, Kai and Gao, Hang and Tan, Juntao and Zhang, Yongfeng},
  booktitle={Advances in Neural Information Processing Systems},
  year={2025},
  doi={10.52202/085713-0593},
  url={https://proceedings.neurips.cc/paper_files/paper/2025/hash/19909c36f51abc4856b4560aff3d36d6-Abstract-Conference.html}
}

@inproceedings{faysse2024colpali,
  title={{ColPali}: Efficient Document Retrieval with Vision Language Models},
  author={Faysse, Manuel and Sibille, Hugues and Wu, Tony and Omrani, Bilel and Viaud, Gautier and Hudelot, C{\'e}line and Colombo, Pierre},
  booktitle={International Conference on Learning Representations},
  year={2025},
  url={https://proceedings.iclr.cc/paper_files/paper/2025/hash/99e9e141aafc314f76b0ca3dd66898b3-Abstract-Conference.html}
}

@inproceedings{yu2024visrag,
  title={{VisRAG}: Vision-based Retrieval-augmented Generation on Multi-modality Documents},
  author={Yu, Shi and Tang, Chaoyue and Xu, Bokai and Cui, Junbo and Ran, Junhao and Yan, Yukun and Liu, Zhenghao and Wang, Shuo and Han, Xu and Liu, Zhiyuan and Sun, Maosong},
  booktitle={International Conference on Learning Representations},
  year={2025},
  url={https://openreview.net/forum?id=zG459X3Xge}
}

@article{zhang2025survey,
  title={A survey on the memory mechanism of large language model-based agents},
  author={Zhang, Zeyu and Dai, Quanyu and Bo, Xiaohe and Ma, Chen and Li, Rui and Chen, Xu and Zhu, Jieming and Dong, Zhenhua and Wen, Ji-Rong},
  journal={ACM Transactions on Information Systems},
  volume={43},
  number={6},
  pages={1--47},
  year={2025},
  publisher={ACM New York, NY},
  doi={10.1145/3748302},
  url={https://dl.acm.org/doi/10.1145/3748302}
}

@inproceedings{asai2024self,
  title={{Self-RAG}: Learning to Retrieve, Generate, and Critique through Self-Reflection},
  author={Asai, Akari and Wu, Zeqiu and Wang, Yizhong and Sil, Avirup and Hajishirzi, Hannaneh},
  booktitle={International Conference on Learning Representations},
  year={2024},
  url={https://openreview.net/forum?id=hSyW5go0v8}
}

@inproceedings{anwar2025remembr,
  title={{ReMEmbR}: Building and reasoning over long-horizon spatio-temporal memory for robot navigation},
  author={Anwar, Abrar and Welsh, John and Biswas, Joydeep and Pouya, Soha and Chang, Yan},
  booktitle={IEEE International Conference on Robotics and Automation},
  year={2025}
}

@inproceedings{gu2024conceptgraphs,
  title={{ConceptGraphs}: Open-vocabulary {3D} scene graphs for perception and planning},
  author={Gu, Qiao and Kuwajerwala, Ali and Morin, Sacha and Jatavallabhula, Krishna Murthy and Sen, Bipasha and Agarwal, Aditya and Rivera, Corban and Paul, William and Ellis, Kirsty and Chellappa, Rama and others},
  booktitle={IEEE International Conference on Robotics and Automation},
  year={2024}
}

@inproceedings{gorlo2026describe,
  title={Describe Anything Anywhere At Any Moment},
  author={Gorlo, Nicolas and Schmid, Lukas and Carlone, Luca},
  booktitle={IEEE/CVF Computer Vision and Pattern Recognition Conference},
  year={2026},
  url={https://openaccess.thecvf.com/content/CVPR2026/html/Gorlo_Describe_Anything_Anywhere_At_Any_Moment_CVPR_2026_paper.html}
}

@inproceedings{fang2019scenememory,
  title={Scene Memory Transformer for Embodied Agents in Long-Horizon Tasks},
  author={Fang, Kuan and Toshev, Alexander and Fei-Fei, Li and Savarese, Silvio},
  booktitle={IEEE/CVF Computer Vision and Pattern Recognition Conference},
  year={2019},
  doi={10.1109/CVPR.2019.00063}
}

@article{xie2024embodiedrag,
  title={{Embodied-RAG}: General Non-parametric Embodied Memory for Retrieval and Generation},
  author={Xie, Quanting and Min, So Yeon and Ji, Pengliang and Yang, Yue and Zhang, Tianyi and Xu, Kedi and Bajaj, Aarav and Salakhutdinov, Ruslan and Johnson-Roberson, Matthew and Bisk, Yonatan},
  journal={arXiv preprint arXiv:2409.18313},
  year={2024},
  doi={10.48550/arXiv.2409.18313},
  url={https://arxiv.org/abs/2409.18313}
}

@article{douglas1973algorithms,
  title={Algorithms for the Reduction of the Number of Points Required to Represent a Digitized Line or its Caricature},
  author={Douglas, David H. and Peucker, Thomas K.},
  journal={Cartographica: The International Journal for Geographic Information and Geovisualization},
  volume={10},
  number={2},
  pages={112--122},
  year={1973},
  doi={10.3138/FM57-6770-U75U-7727}
}

@inproceedings{wang2025videotree,
  title={{VideoTree}: Adaptive Tree-based Video Representation for {LLM} Reasoning on Long Videos},
  author={Wang, Ziyang and Yu, Shoubin and Stengel-Eskin, Elias and Yoon, Jaehong and Cheng, Feng and Bertasius, Gedas and Bansal, Mohit},
  booktitle={IEEE/CVF Computer Vision and Pattern Recognition Conference},
  year={2025}
}

@article{zou2026taskmem,
  title={Task-Focused Memorization for Multimodal Agents},
  author={Zou, Tao and He, Yichen and Qiu, Tian and Lin, Yuan and Li, Hang},
  journal={arXiv preprint arXiv:2605.31075},
  year={2026}
}

@inproceedings{chen2024spatialvlm,
  title={{SpatialVLM}: Endowing Vision-Language Models with Spatial Reasoning Capabilities},
  author={Chen, Boyuan and Xu, Zhuo and Kirmani, Sean and Ichter, Brian and Sadigh, Dorsa and Guibas, Leonidas and Xia, Fei},
  booktitle={IEEE/CVF Computer Vision and Pattern Recognition Conference},
  year={2024},
  url={https://openaccess.thecvf.com/content/CVPR2024/html/Chen_SpatialVLM_Endowing_Vision-Language_Models_with_Spatial_Reasoning_Capabilities_CVPR_2024_paper.html}
}

@inproceedings{cheng2024spatialrgpt,
  title={{SpatialRGPT}: Grounded Spatial Reasoning in Vision-Language Models},
  author={Cheng, An-Chieh and Yin, Hongxu and Fu, Yang and Guo, Qiushan and Yang, Ruihan and Kautz, Jan and Wang, Xiaolong and Liu, Sifei},
  booktitle={Advances in Neural Information Processing Systems},
  year={2024},
  doi={10.52202/079017-4293},
  url={https://proceedings.neurips.cc/paper_files/paper/2024/hash/f38cb4cf9a5eaa92b3cfa481832719c6-Abstract-Conference.html}
}

@misc{qwen3.5,
  title = {{Qwen3.5}: Towards Native Multimodal Agents},
  author = {{Qwen Team}},
  month = {February},
  year = {2026},
  url = {https://qwen.ai/blog?id=qwen3.5}
}

@inproceedings{werby2024hovsg,
  title={Hierarchical Open-Vocabulary {3D} Scene Graphs for Language-Grounded Robot Navigation},
  author={Werby, Abdelrhman and Huang, Chenguang and B{\"u}chner, Martin and Valada, Abhinav and Burgard, Wolfram},
  booktitle={Robotics: Science and Systems},
  year={2024},
  doi={10.15607/RSS.2024.XX.077},
  url={https://www.roboticsproceedings.org/rss20/p077.html}
}

@article{maggio2024clio,
  title={{Clio}: Real-Time Task-Driven Open-Set {3D} Scene Graphs},
  author={Maggio, Dominic and Chang, Yun and Hughes, Nathan and Trang, Matthew and Griffith, Dan and Dougherty, Carlyn and Cristofalo, Eric and Schmid, Lukas and Carlone, Luca},
  journal={IEEE Robotics and Automation Letters},
  volume={9},
  number={10},
  pages={8921--8928},
  year={2024},
  doi={10.1109/LRA.2024.3451395},
  url={https://ieeexplore.ieee.org/document/10659066/}
}

@inproceedings{hu2025threedllmmem,
  title={{3DLLM-Mem}: Long-Term Spatial-Temporal Memory for Embodied {3D} Large Language Model},
  author={Hu, Wenbo and Hong, Yining and Wang, Yanjun and Gao, Leison and Wei, Zibu and Yao, Xingcheng and Peng, Nanyun and Bitton, Yonatan and Szpektor, Idan and Chang, Kai-Wei},
  booktitle={Advances in Neural Information Processing Systems},
  year={2025},
  doi={10.52202/085713-2278},
  url={https://proceedings.neurips.cc/paper_files/paper/2025/hash/61f527a737e4ba61f3e10d6c3f0c4b55-Abstract-Conference.html}
}

@inproceedings{saxena2025grapheqa,
  title={{GraphEQA}: Using {3D} Semantic Scene Graphs for Real-time Embodied Question Answering},
  author={Saxena, Saumya and Buchanan, Blake and Paxton, Chris and Liu, Peiqi and Chen, Bingqing and Vaskevicius, Narunas and Palmieri, Luigi and Francis, Jonathan and Kroemer, Oliver},
  booktitle={Conference on Robot Learning},
  year={2025},
  url={https://proceedings.mlr.press/v305/saxena25a.html}
}

@inproceedings{ginting2025mindpalace,
  title={Enter the Mind Palace: Reasoning and Planning for Long-term Active Embodied Question Answering},
  author={Ginting, Muhammad Fadhil and Kim, Dong-Ki and Meng, Xiangyun and Reinke, Andrzej Marek and Krishna, Bandi Jai and Kayhani, Navid and Peltzer, Oriana and Fan, David and Shaban, Amirreza and Kim, Sung-Kyun and Kochenderfer, Mykel and Agha-mohammadi, Ali-akbar and Omidshafiei, Shayegan},
  booktitle={Conference on Robot Learning},
  year={2025},
  url={https://proceedings.mlr.press/v305/ginting25a.html}
}

@inproceedings{ma2025spatialllm,
  title={{SpatialLLM}: A Compound {3D}-Informed Design towards Spatially-Intelligent Large Multimodal Models},
  author={Ma, Wufei and Ye, Luoxin and de Melo, Celso M and Yuille, Alan and Chen, Jieneng},
  booktitle={IEEE/CVF Computer Vision and Pattern Recognition Conference},
  year={2025},
  url={https://openaccess.thecvf.com/content/CVPR2025/html/Ma_SpatialLLM_A_Compound_3D-Informed_Design_towards_Spatially-Intelligent_Large_Multimodal_Models_CVPR_2025_paper.html}
}

@inproceedings{zhu2025llava3d,
  title={{LLaVA-3D}: A Simple yet Effective Pathway to Empowering {LMMs} with {3D} Capabilities},
  author={Zhu, Chenming and Wang, Tai and Zhang, Wenwei and Pang, Jiangmiao and Liu, Xihui},
  booktitle={International Conference on Computer Vision},
  year={2025},
  url={https://openaccess.thecvf.com/content/ICCV2025/html/Zhu_LLaVA-3D_A_Simple_yet_Effective_Pathway_to_Empowering_LMMs_with_ICCV_2025_paper.html}
}

@inproceedings{wang2024videoagent,
  title={{VideoAgent}: Long-Form Video Understanding with Large Language Model as Agent},
  author={Wang, Xiaohan and Zhang, Yuhui and Zohar, Orr and Yeung-Levy, Serena},
  booktitle={European Conference on Computer Vision},
  year={2024},
  doi={10.1007/978-3-031-72989-8_4},
  url={https://www.ecva.net/papers/eccv_2024/papers_ECCV/html/10325_ECCV_2024_paper.php}
}

@inproceedings{ye2025temporalsearch,
  title={Re-thinking Temporal Search for Long-Form Video Understanding},
  author={Ye, Jinhui and Wang, Zihan and Sun, Haosen and Chandrasegaran, Keshigeyan and Durante, Zane and Eyzaguirre, Cristobal and Bisk, Yonatan and Niebles, Juan Carlos and Adeli, Ehsan and Fei-Fei, Li and Wu, Jiajun and Li, Manling},
  booktitle={IEEE/CVF Computer Vision and Pattern Recognition Conference},
  year={2025},
  url={https://openaccess.thecvf.com/content/CVPR2025/html/Ye_Re-thinking_Temporal_Search_for_Long-Form_Video_Understanding_CVPR_2025_paper.html}
}

@inproceedings{lee2024readagent,
  title={A Human-Inspired Reading Agent with Gist Memory of Very Long Contexts},
  author={Lee, Kuang-Huei and Chen, Xinyun and Furuta, Hiroki and Canny, John and Fischer, Ian},
  booktitle={International Conference on Machine Learning},
  year={2024},
  url={https://proceedings.mlr.press/v235/lee24c.html}
}

@inproceedings{gutierrez2024hipporag,
  title={{HippoRAG}: Neurobiologically Inspired Long-Term Memory for Large Language Models},
  author={Guti{\'e}rrez, Bernal Jim{\'e}nez and Shu, Yiheng and Gu, Yu and Yasunaga, Michihiro and Su, Yu},
  booktitle={Advances in Neural Information Processing Systems},
  year={2024},
  doi={10.52202/079017-1902},
  url={https://proceedings.neurips.cc/paper_files/paper/2024/hash/6ddc001d07ca4f319af96a3024f6dbd1-Abstract-Conference.html}
}

@inproceedings{gutierrez2025hipporag2,
  title={From {RAG} to Memory: Non-Parametric Continual Learning for Large Language Models},
  author={Guti{\'e}rrez, Bernal Jim{\'e}nez and Shu, Yiheng and Qi, Weijian and Zhou, Sizhe and Su, Yu},
  booktitle={International Conference on Machine Learning},
  year={2025},
  url={https://proceedings.mlr.press/v267/gutierrez25a.html}
}

@inproceedings{kang2025memoryos,
  title={Memory {OS} of {AI} Agent},
  author={Kang, Jiazheng and Ji, Mingming and Zhao, Zhe and Bai, Ting},
  booktitle={Conference on Empirical Methods in Natural Language Processing},
  year={2025},
  doi={10.18653/v1/2025.emnlp-main.1318},
  url={https://aclanthology.org/2025.emnlp-main.1318/}
}

@inproceedings{jin2025searchr1,
  title={{Search-R1}: Training {LLM}s to Reason and Leverage Search Engines with Reinforcement Learning},
  author={Jin, Bowen and Zeng, Hansi and Yue, Zhenrui and Yoon, Jinsung and Arik, Sercan O. and Wang, Dong and Zamani, Hamed and Han, Jiawei},
  booktitle={Conference on Language Modeling},
  year={2025},
  url={https://openreview.net/forum?id=Rwhi91ideu}
}

@inproceedings{zhu2026hymem,
  title={Hybrid Self-evolving Structured Memory for Computer-Use Agents},
  author={Zhu, Sibo and Wu, Wenyi and Zhou, Kun and Wang, Stephen and Huang, Biwei},
  booktitle={Findings of the Association for Computational Linguistics: ACL 2026},
  year={2026},
  doi={10.18653/v1/2026.findings-acl.549},
  url={https://aclanthology.org/2026.findings-acl.549/}
}

@inproceedings{gao2026scaling,
  title={Scaling Beyond Context: A Survey of Multimodal Retrieval-Augmented Generation for Document Understanding},
  author={Gao, Sensen and Zhao, Shanshan and Jiang, Xu and Duan, Lunhao and Chng, Yong Xien and Chen, Qing-Guo and Luo, Weihua and Zhang, Kaifu and Bian, Jia-Wang and Gong, Mingming},
  booktitle={Annual Meeting of the Association for Computational Linguistics},
  year={2026},
  doi={10.18653/v1/2026.acl-long.204},
  url={https://aclanthology.org/2026.acl-long.204/}
}

\end{document}